\documentclass[pdflatex,bst/sn-mathphys-num,iicol]{sn-jnl}%

\usepackage{graphicx}%
\usepackage{multirow}%
\usepackage{amsmath,amssymb,amsfonts}%
\usepackage{amsthm}%
\usepackage{mathrsfs}%
\usepackage[title]{appendix}%
\usepackage{xcolor}%
\usepackage{textcomp}%
\usepackage{manyfoot}%
\usepackage{booktabs}%
\usepackage{algorithm}%
\usepackage{algorithmicx}%
\usepackage{algpseudocode}%
\usepackage{listings}%
\usepackage{natbib}

\usepackage{booktabs}
\usepackage{tabularx}
\usepackage{epsfig}
\usepackage{colortbl}
\usepackage{color}
\usepackage{xcolor}
\usepackage{setspace}
\usepackage[normalem]{ulem}
\usepackage{tikz}
\usepackage{comment}
\usepackage{float}
\usepackage{caption}
\usepackage{adjustbox}
\usepackage{pifont}%
\usepackage{wrapfig}
\newcommand{\cmark}{\ding{51}}%
\newcommand{\xmark}{\ding{55}}%

\definecolor{top1}{RGB}{245,152,153}
\definecolor{top2}{RGB}{253,205,154}
\definecolor{top3}{RGB}{248,244,140}

\definecolor{blgray}{gray}{0.97}
\definecolor{mygray}{gray}{.93}
\definecolor{nicegreen}{rgb}{0.1, 0.6, 0.2}
\newcommand{\Revision}[1]{{#1}}

\usepackage[capitalize]{cleveref}
\crefname{section}{Sec.}{Secs.}
\Crefname{section}{Section}{Sections}
\Crefname{table}{Table}{Tables}
\crefname{table}{Tab.}{Tabs.}

\theoremstyle{thmstyleone}%

\theoremstyle{thmstyletwo}%

\theoremstyle{thmstylethree}%

\begin{document}

\title[Article Title]{
High-Quality Sound Separation Across Diverse Categories via Visually-Guided Generative Modeling
}

\author[1]{\fnm{Chao} \sur{Huang}}\email{chuang65@cs.rochester.edu}
\author[1]{\fnm{Susan} \sur{Liang}}\email{sliang22@ur.rochester.edu}
\author[2]{\fnm{Yapeng} \sur{Tian}}\email{yapeng.tian@utdallas.edu}
\author[3]{\fnm{Anurag} \sur{Kumar}}\email{anuragkr90@meta.com}
\author[1]{\fnm{Chenliang} \sur{Xu}}\email{chenliang.xu@rochester.edu}

\affil[1]{\orgdiv{Department of Computer Science}, \orgname{University of Rochester}, \orgaddress{\city{Rochester}, \postcode{14627}, \state{NY}, \country{USA}}}
\affil[2]{\orgdiv{Department of Computer Science}, \orgname{The University of Texas
at Dallas}, \orgaddress{\city{Richardson}, \postcode{75080}, \state{TX}, \country{USA}}}
\affil[3]{\orgdiv{Meta Reality Labs Research}, \orgaddress{\city{Redmond}, \postcode{98052}, \state{WA}, \country{USA}}}

\abstract{

 We propose DAVIS, a \textbf{D}iffusion-based \textbf{A}udio-\textbf{VI}sual \textbf{S}eparation framework that solves the audio-visual sound source separation task through generative learning. Existing methods typically frame sound separation as a mask-based regression problem, achieving significant progress.  However, they face limitations in capturing the complex data distribution required for high-quality separation of sounds from diverse categories. In contrast, DAVIS circumvents these issues by leveraging potent generative modeling paradigms, specifically Denoising Diffusion Probabilistic Models (DDPM) and the more recent Flow Matching (FM), integrated within a specialized Separation U-Net architecture. Our framework operates by synthesizing the desired separated sound spectrograms directly from a noise distribution, conditioned concurrently on the mixed audio input and associated visual information. The inherent nature of its generative objective makes DAVIS particularly adept at producing high-quality sound separations for diverse sound categories. We present comparative evaluations of DAVIS, encompassing both its DDPM and Flow Matching variants, against leading methods on the standard AVE and MUSIC datasets. The results affirm that both variants surpass existing approaches in separation quality, highlighting the efficacy of our generative framework for tackling the audio-visual source separation task. Our code is available here: \url{https://github.com/WikiChao/DAVIS}.

}


\maketitle

\section{Introduction}\label{intro}

\begin{figure*}[!tbp]
    \centering
    \includegraphics[width=1\textwidth]{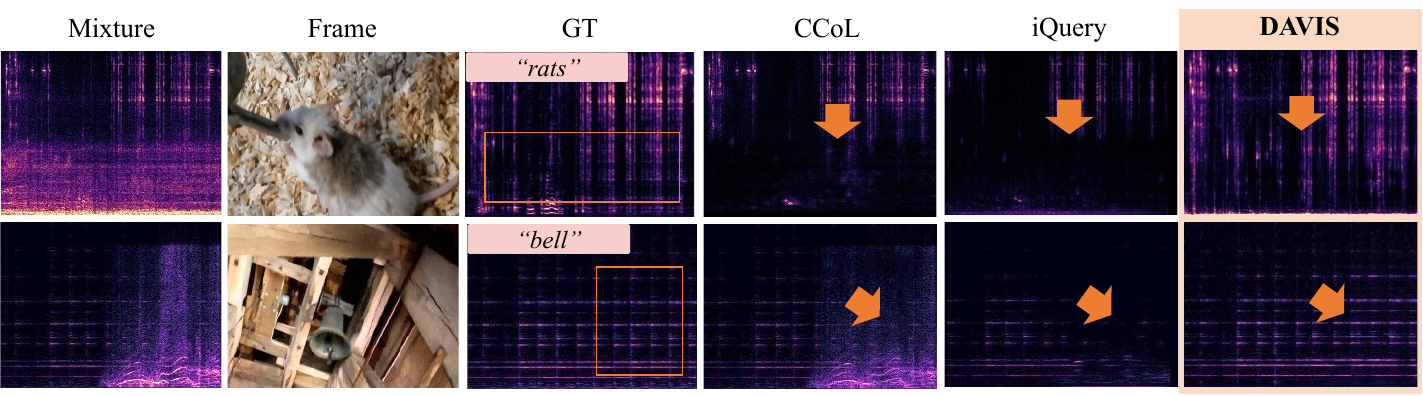}
    \caption{
    \textbf{Separation results on diverse time-frequency structures} are shown for SOTA discriminative methods and our proposed generative framework (DAVIS shown). Each row displays the audio mixture, reference visual frame, ground truth magnitude, and predicted magnitudes from DAVIS, iQuery~\citep{chen2023iquery}, and CCoL~\citep{tian2021cyclic}. Our generative approach successfully recovers suppressed time-frequency structures (highlighted in the box), where mask-regression methods often fail. Similar improvements are observed with DAVIS-Flow (see \cref{fig:vis}).
    }
    \label{fig:motivation}
\end{figure*}

Visually-guided sound source separation, also referred to as audio-visual separation,  is crucial for gauging a machine's ability to process multisensory information~\citep{zhao2018sound,gao2018learning}. It aims to separate individual sounds from a complex audio mixture by using visual cues about the objects that are producing the sounds, \textit{e.g.}, separate the ``barking'' sound from the mixture by querying the ``dog'' object. To be effective, separation models should achieve \textbf{\textit{high-quality}} results across a \textbf{\textit{diverse}} range of sound categories, yielding a realistic auditory experience. The community has devoted considerable effort to tackling this task~\citep{zhao2018sound,gao2019co,gan2020music,chatterjee2021visual,tian2021cyclic,dong2023clipsep,zhu2022visually,chen2023iquery}, focusing on aspects like advanced separation frameworks~\citep{zhao2018sound, gao2019co, chatterjee2021visual, chen2023iquery}, novel training piplines~\citep{tian2021cyclic}, and leveraging auxiliary visual information~\citep{gan2020music} for better performance. Predominant methods have often relied on discriminative objectives, such as regressing time-frequency masks~\citep{zhao2018sound} or directly reconstructing spectrograms~\citep{owens2018audio}.

While these methods have yielded promising results in separation performance, they can struggle particularly with the nuances of highly diverse time-frequency structures and challenging conditions where sounds are intricately combined or the target sound is heavily obscured (as illustrated by examples in \cref{fig:motivation}). These situations pose significant obstacles for approaches that rely on directly regressing masks or spectrograms. Hence, this naturally prompts the question: \textit{could alternative modeling paradigms be better equipped to capture the complexities of data distributions, leverage precise audio-visual associations more effectively, and generate separated sounds of higher quality?}

We address this question by investigating the potential of powerful generative modeling frameworks for audio-visual separation. One prominent class, Denoising Diffusion Probabilistic Models (DDPMs)~\citep{ho2020denoising,nichol2021improved,song2020denoising}, has demonstrated remarkable capabilities in synthesizing diverse and high-quality data, including images~\citep{dhariwal2021diffusion} and audio~\citep{kong2020diffwave}. More recently, Flow Matching (FM) models~\citep{lipman2022flow, liu2022flow} have emerged as a compelling alternative paradigm, potentially offering advantages such as simpler training procedures due to the ``straightness'' of their induced trajectories and efficient inference through continuous-time flows defined by Ordinary Differential Equations (ODEs). Nevertheless, successfully adapting either DDPMs or FM to the specific requirements of audio-visual separation presents non-trivial challenges. These primarily stem from the necessity of designing specialized architectures capable of capturing complex audio-visual correspondences while accurately modeling the distinct characteristics of magnitude spectrograms.

In this paper, we introduce DAVIS, a novel generative framework specifically designed for audio-visual separation. Moving away from conventional mask regression techniques, our framework approaches separation as a conditional generation process. DAVIS operates by iteratively synthesizing the target magnitude spectrogram starting from Gaussian noise, guided by both the audio mixture and relevant visual information. We design the DAVIS framework through two distinct instantiations based on powerful generative paradigms: the original DAVIS employs the DDPM approach, while DAVIS-Flow utilizes Flow Matching. As detailed in \cref{subsubsec:fm}, these two pipelines are closely related, fundamentally representing different perspectives on generative modeling that start from a simple distribution. This generative strategy effectively distributes the challenge of recovering complex time-frequency patterns, making the framework adaptable to a variety of scenarios, including difficult cases like those depicted in \cref{fig:motivation}. A critical shared element within both the DAVIS and DAVIS-Flow implementations is our proposed Separation U-Net architecture. This network is specifically engineered to handle the complexities of spectrogram data and the intricate nature of audio-visual correlations. To capture important long-range dependencies often present in spectrograms, we integrate Convolution-Attention (CA) blocks into the Separation U-Net, enabling it to process both local features and non-local contexts. Furthermore, to improve the learning and utilization of audio-visual associations, we designed a Feature Interaction Module (FIM) for the effective injection and processing of visual cues.

A further challenge encountered in this task is the prevalence of silent regions (corresponding to near-zero values) within magnitude spectrograms, which results in a significantly skewed data distribution. To mitigate this, we adopt a more robust $\mathcal{L}_1$ loss for training both the DDPM and FM variants of our framework, diverging from the commonly used $\mathcal{L}_2$ loss. Moreover, these silent areas are not merely data anomalies but can offer valuable structural guidance during the inference phase, indicating where the target sound is absent within the mixture. We capitalize on this by introducing a silence mask-guided inference strategy. This approach is applicable to both the iterative sampling procedure of DAVIS (DDPM) and the ODE solving process of DAVIS-Flow (FM), allowing us to refine the generated spectrograms and enhance their consistency with the input mixture.

Experimental evaluation conducted on the AVE~\citep{tian2018audio} and MUSIC~\citep{zhao2018sound} datasets demonstrates the effectiveness of our proposed generative framework, with both the DAVIS (DDPM-based) and DAVIS-Flow (FM-based) instantiations consistently achieving superior separation quality compared to existing state-of-the-art methods. Furthermore, by leveraging the rich image-text embedding space provided by the CLIP model~\citep{radford2021learning}, our framework unlocks a novel capability: zero-shot text-guided audio separation, which is achieved by transferring the audio-visual correlations learned by our model into the textual domain.
Our contributions are summarized as follows:
\begin{itemize}
    \item We introduce a novel generative framework for audio-visual separation by successfully applying and demonstrating the effectiveness of two advanced generative modeling paradigms: Denoising Diffusion Probabilistic Models (DDPMs) and Flow Matching (FM).
    \item We design and propose a specialized Separation U-Net architecture, shared across both the DDPM and FM instantiations, equipped with Convolution-Attention blocks for capturing multi-scale dependencies and an Audio-Visual Feature Interaction Module for effective multimodal integration.
    \item  We identify and address the challenge of skewed data distributions in magnitude spectrograms by using a robust  $\mathcal{L}_1$ loss for training and proposing a novel silence mask-guided inference strategy applicable to both the iterative DDPM sampling and FM's ODE solving process.
    \item We empirically validate our generative approach through extensive experiments on standard datasets (AVE and MUSIC), showing that both the DAVIS and DAVIS-Flow methods achieve performance that is competitive with or superior to existing discriminative techniques.
\end{itemize}
This manuscript is a significant extension of a preliminary version that was accepted to ACCV 2024~\citep{Huang_2024_ACCV}, which received the Best Paper Honorable Mention award. In this work, we introduce substantial new content and analysis. The key additions include: (i) the introduction of an improved method, DAVIS-Flow, which applies the recent Flow Matching framework to the audio-visual separation task (\cref{subsubsec:fm}); (ii) a detailed discussion on the relationships and distinctions between the original DAVIS (DDPM-based) and DAVIS-Flow (FM-based) variants, illustrating how Flow Matching is integrated into our framework (\cref{subsubsec:fm}); (iii) a thorough evaluation of the DAVIS-Flow method on the AVE and MUSIC datasets to demonstrate its effectiveness in audio-visual separation (\cref{subsec:comparison}); (iv) supplementary analysis of the core DAVIS framework, including investigations into loss functions, the effectiveness of visual condition aggregation, and the impact of the number of sampling steps (\cref{subsec:analysis}); and (v) a comparative analysis providing detailed insights into the performance and efficiency differences between DAVIS and DAVIS-Flow, particularly regarding training and inference speed (\cref{subsec:versu}).

\section{Related Work}
\noindent\textbf{Audio-Visual Sound Source Separation.}
This section reviews recent advancements in audio-visual sound source separation, building upon earlier research efforts in signal processing-based sound separation~\citep{virtanen2007monaural,smaragdis2003non}. Contemporary deep learning-based approaches have been successfully applied to separating a range of audio types, including speech~\citep{ephrat2018looking,owens2018audio,afouras2020self,michelsanti2021overview}, musical instruments~\citep{zhao2018sound,gan2020music,tian2021cyclic,gao2019co,zhao2019sound,chatterjee2021visual,tan2023language}, and general sound events~\citep{gao2018learning,mittal2022learning,tzinis2020into,tzinis2022audioscopev2,chatterjee2022learning,zhu2022visually,dong2023clipsep,chen2023iquery}. A common training methodology involves mixing audio streams from different videos to generate supervised training examples. A neural network, frequently based on a U-Net architecture, is then trained to regress a separation mask conditioned on associated visual features~\citep{zhao2018sound}. Recent research trends encompass both domain-specific and open-domain sound source separation challenges~\citep{tzinis2020into,mittal2022learning,zhu2022visually,dong2023clipsep,chen2023iquery,huang2025zerosep}. Many existing methods, however, rely on auxiliary information such as text queries~\citep{dong2023clipsep,huang2025learning}, pose data~\citep{huang2024modeling}, motion cues~\citep{mittal2022learning,zhu2022visually}, or class labels~\citep{chen2023iquery} to achieve satisfactory performance. In contrast, this work introduces a novel generative audio-visual separation framework that demonstrates competitive or superior performance for separating both specific and open-domain sound sources without requiring such additional cues beyond the visual input and audio mixture.

\noindent\textbf{Diffusion Models.}
Diffusion models, including Denoising Diffusion Probabilistic Models (DDPMs)~\citep{ho2020denoising,song2020score,song2019generative}, represent a class of deep generative models that operate by gradually transforming a sample from a simple random distribution (typically Gaussian noise) into a data sample through an iterative denoising process. In recent years, diffusion models have achieved impressive results across diverse domains, including image synthesis~\citep{dhariwal2021diffusion,avrahami2022blended,ramesh2022hierarchical,gu2022vector,nichol2021glide,ho2022video,singer2022make,ruiz2022dreambooth,saharia2022photorealistic,huang2025fresca,huang2024scaling}, natural language processing applications~\citep{austin2021structured,gong2022diffuseq,li2022diffusion,chen2022analog}, audio generation and synthesis~\citep{kong2020diffwave,popov2021grad,lee2021nu,chen2022analog,chen2020wavegrad,huang2022prodiff,scheibler2023diffusion}, and multimodal content generation involving audio-visual data~\citep{ruan2022mm,liang2024language}. Beyond their generative capabilities, there is growing interest in applying diffusion models to discriminative tasks. Initial studies have explored their use in image segmentation~\citep{amit2021segdiff,baranchuk2021label,brempong2022denoising} and object detection~\citep{chen2022diffusiondet}. Despite this expanding interest, successful applications of generative diffusion models specifically to audio-visual scene understanding tasks remain relatively limited. A few recent works have investigated diffusion-based approaches for audio-visual speech separation and enhancement~\citep{lee2023seeing,chou2023av2wav}. However, these efforts have typically focused narrowly on speech, rather than the more generalized and challenging problem of audio-visual sound separation. Key challenges in applying diffusion models to this domain include the design of effective network architectures capable of capturing intricate audio-visual correspondences and adapting diffusion models to handle the unique statistical properties of audio data representations like spectrograms. This paper contributes to addressing this gap by presenting a diffusion-based model for broader audio-visual sound separation, enabled by a novel separation architecture designed to effectively learn the complex relationships between audio and visual modalities.

\noindent\textbf{Flow Matching Models.}
Flow Matching (FM) is a recently developed class of generative models that learn to transport a simple base distribution (e.g., Gaussian noise) to a complex target data distribution by training a neural network to predict a vector field~\citep{lipman2022flow, liu2022flow}. Unlike Diffusion Models which typically involve learning the score function of data distributions perturbed by a stochastic process, Flow Matching directly learns a conditional vector field that defines a deterministic Ordinary Differential Equation (ODE). This ODE can then be solved to generate samples by starting from the base distribution and integrating the learned vector field. A key advantage highlighted in recent work is the simplicity and stability of training the vector field prediction~\citep{lipman2022flow}. Furthermore, sampling from trained Flow Matching models can be more efficient than iterative denoising in DDPMs, requiring only a single pass through an ODE solver (though potentially with multiple evaluation steps). While Flow Matching shares conceptual links with diffusion models, especially through the lens of continuous-time generative models and their connection to probability flows~\citep{gao2025diffusionmeetsflow}, it offers an alternative modeling and inference paradigm. Originally applied to image generation, FM has shown promise in various domains \citep{mehta2024matcha,du2024cosyvoice,polyak2024movie,jin2024pyramidal,liu2023generative,gat2024discrete,liangbinauralflow,liangpiavas}. Our work explores the effectiveness of adapting the Flow Matching paradigm for the complex task of audio-visual sound source separation, presenting DAVIS-Flow as a novel application of this generative modeling technique to the audio-visual domain.

\begin{figure*}[t]
\begin{center}
\includegraphics[width=0.98\linewidth]{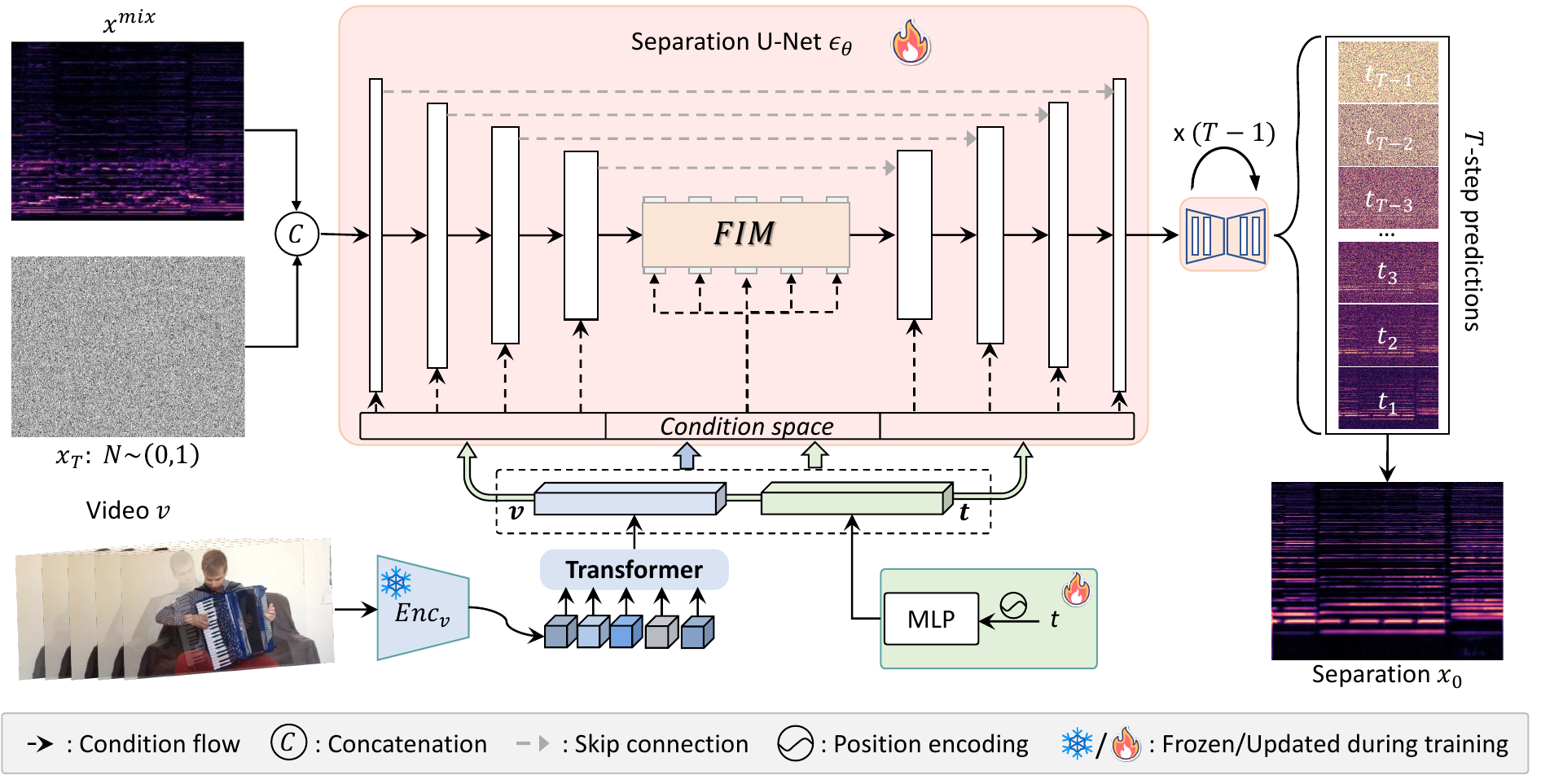}
\end{center}
  \caption{\textbf{Overview of the DAVIS framework.} 
We aim to synthesize $x_0$ from the mixture $x^{mix}$, visual stream $v$, and timestep $t$. Starting with $x_T$ from a standard distribution, we encode $v=\{I_j\}_{j=1}^K$ and $t$ into the embedding space. A temporal transformer generates the visual feature $\boldsymbol{v}$, which, along with $\boldsymbol{t}$, conditions the Separation U-Net $\epsilon_\theta$ to iteratively denoise $x_T$ into $x_0$. $\boldsymbol{v}$ is used only in the Feature Interaction Module for audio-visual association, while $\boldsymbol{t}$ is used throughout. For the DDPM variant (DAVIS), the U-Net $\epsilon_\theta$ is trained to estimate the noise added at timestep $t$ to iteratively denoise $x_T$ to $x_0$. For the Flow Matching variant (DAVIS-Flow), the same U-Net $v_\theta$ is trained to approximate the vector field defining a continuous-time flow that transports $x_T$ to $x_0$. The initial input $x_T$ is sampled from a standard distribution and the conditioning inputs ($x^{\text{mix}}$, $v$, $t$) remain consistent for both variants.
  }
\label{fig: framework}
\end{figure*}

\section{Method}
\label{sec:method}
In this section, we introduce DAVIS, our novel audio-visual separation framework based on generative modeling. We present two variants: the original DAVIS, which utilizes Denoising Diffusion Probabilistic Models (DDPM)~\citep{ho2020denoising}, and DAVIS-Flow, which leverages the Flow Matching (FM) paradigm~\citep{lipman2022flow, liu2022flow}. We begin by providing a brief recap of both DDPM and FM in \cref{subsec:preliminaries}. Then, we describe our task setup and give a method overview in \cref{subsec:setup}, highlighting how both paradigms are employed. Next, we present our proposed Separation U-Net, the core shared architecture for both variants, in \cref{subsec:network}. Furthermore, we discuss the distinct training pipelines for DAVIS and DAVIS-Flow in \cref{subsec:training}. Finally, we introduce the silence mask-guided inference strategy applicable to both approaches in \cref{subsec:inference}.

\subsection{Preliminaries: Generative Modeling Paradigms}
\label{subsec:preliminaries}

\subsubsection{Denoising Diffusion Probabilistic Models (DDPM)}
\label{subsubsec:ddpm}
DDPMs~\citep{ho2020denoising} consist of a forward and a reverse process. The forward process is a fixed Markov chain that gradually adds Gaussian noise to a data sample $x_0$ over $T$ timesteps according to a variance schedule $\beta_1, ..., \beta_T$. Sampling $x_t$ at an arbitrary timestep $t$ is done via:
\begin{equation}
    q(x_t | x_0) = \mathcal{N}(x_t; \sqrt{\Bar{\alpha}_t}x_{0}, (1-\Bar{\alpha}_t)\mathbf{I}),
\label{eq:ddpm_forward}
\end{equation}
where $\alpha_t = 1 - \beta_t$ and $\Bar{\alpha}_t = \prod_{s=1}^t \alpha_s$. As $t \rightarrow T$, $x_t$ approaches a standard Gaussian distribution $\mathcal{N}(\mathbf{0}, \mathbf{I})$. Specifically, $x_t$ can be derived through:
\begin{equation}
    x_t = \sqrt{\Bar{\alpha}_t}x_0 +\sqrt{1-\Bar{\alpha}_t}\epsilon, \quad \epsilon \sim \mathcal{N}(\mathbf{0}, \mathbf{I}). 
\label{eq:ddpm_forward}
\end{equation}

The reverse process learns to denoise $x_t$ back to $x_0$ using a parameterized Markov chain:
\begin{equation}
    p_\theta(x_{0:T}) = p(x_{T})\prod_{t=1}^T p_\theta(x_{t-1}|x_t),
\label{eq:ddpm_reverse}
\end{equation}
where each transition step $p_\theta(x_{t-1}|x_t) = \mathcal{N}(x_{t-1};\boldsymbol{\mu_\theta}(x_t, t, \boldsymbol{c}), \Tilde{\beta_t}\mathbf{I})$ depends on a neural network $\theta$ that predicts the mean $\boldsymbol{\mu_\theta}$, conditioned on the noisy data $x_t$, timestep $t$, and potentially external context $\boldsymbol{c}$. The variances $\Tilde{\beta_t}$ are typically fixed~\citep{ho2020denoising}. Training often involves optimizing a simplified objective~\citep{ho2020denoising} where the network $\epsilon_\theta$ predicts the noise $\epsilon$ added at step $t$:
\begin{equation}
    \mathcal{L}_{DDPM}(\theta) = \mathbb{E}_{t}[ ||\epsilon - \epsilon_\theta(x_t, t, \boldsymbol{c} )||^2],
\label{eq:ddpm_loss}
\end{equation}
where $t$ is uniformly sampled from $\{1, ..., T\}$.

\subsubsection{Flow Matching (FM)}
\label{subsubsec:fm}
Flow Matching~\citep{lipman2022flow, liu2022flow} offers an alternative generative modeling approach by learning a continuous-time probability path $p_t(x)$ that transforms a simple prior distribution $p_0$ (\textit{e.g.}, Gaussian noise) into a target data distribution $p_1$. This transformation is defined by an Ordinary Differential Equation (ODE): $dx_t/dt = v_\theta(x_t,t)$, where $v_\theta(\cdot)$ is a time-dependent vector field.

Conditional Flow Matching (CFM)~\citep{liu2022flow} extends this to conditional generation. Given a condition $\boldsymbol{c}$, the goal is to learn a conditional vector field $v_\theta(x_t, t, \boldsymbol{c})$ that approximates the target vector field $u_t(x_t|\boldsymbol{c})$ driving the conditional probability path $p_t(x|\boldsymbol{c})$. A common choice for the probability path connecting $x_0 \sim p_0$ and $x_1 \sim p_1(\cdot|\boldsymbol{c})$ is the linear interpolation path: 
\begin{equation}
    x_t = t x_1 + (1-t) x_0, \quad \text{for}\ t \in [0, 1].
\label{eq:fm_forward}
\end{equation}
The corresponding target vector field is $u_t(x_t|x_1) = x_1 - x_0$. Given this, the FM training objective aims to minimize the difference between the predicted and target vector fields:
\begin{equation}
    \mathcal{L}_{FM}(\theta) = \mathbb{E}_{t}[ ||v_\theta(x_t, t, \boldsymbol{c}) - u_t(x_t|x_1)||^2 ],
\label{eq:fm_loss}
\end{equation}
where $x_t$ is sampled from the chosen path defined by $x_0$ and $x_1$ (\textit{e.g.}, the linear path above). Sampling from an FM model involves solving the probability flow ODE $dx_t/dt = v_\theta(x_t, t, \boldsymbol{c})$ from $t=0$ to $t=1$, starting with a sample $x_0 \sim p_0$.

\noindent\textbf{Connection from DDPM to FM.} 
Although DDPM and FM processes are defined somewhat differently, we highlight their deep underlying connection~\citep{gao2025diffusionmeetsflow}, which makes them readily adaptable within our DAVIS framework. This shared foundation allows us to consider them as variations within a unified generative approach. Their commonalities include sharing the same forward process under the assumption that one end of the flow matches a Gaussian distribution and the DDPM utilizes a specific noise schedule. Furthermore, their sampling processes are notably similar, both involving an iterative update mechanism that estimates the clean data at the current timestep. Hence, we position DAVIS and DAVIS-Flow as two variants under the umbrella of a single generative separation framework, rather than as fundamentally separate methods. \cref{subsec:setup} will subsequently demonstrate how the same model architecture serves both variants.

\subsection{Task Setup and Method Overview}
\label{subsec:setup}
Given an unlabeled video clip $V$, we can extract an audio-visual pair $(a, v)$, where $a$ and $v$ are the audio and visual streams, respectively. In real-world scenarios, the audio stream can be a mixture of $N$ individual sound sources, denoted as $a = \sum_{i=1}^N s_i$, where each source $s_i$ can be of various categories. Meanwhile, the visual stream $v$ is typically a synchronized video of $K$ frames, denoted as $v=\{I_j\}_{j=1}^K$. The visually-guided sound source separation task aims to utilize visual cues from $v$ to help separate $a$ into $N$ individual sources $s_i$. Since no labels are provided to distinguish the sound sources $s_i$, prior works~\citep{zhao2018sound,tian2021cyclic,huang2023egocentric} have commonly used a ``mix and separate'' strategy, which involves mixing audio streams from two different videos and manually create the mixture: $a^{mix} = a^{(1)} + a^{(2)}$. In practice, audio is usually transformed into magnitude spectrogram by short-time Fourier transform $x = \mathbf{STFT}(a) \in \mathbb{R}^{ T \times F}$, allowing for manipulations in the 2D Time-Frequency domain. Here, $F$ and $T$ are the numbers of frequency bins and time frames, respectively. Consequently, the goal of training is to learn a separation network capable of mapping $\boldsymbol{f: (x^{mix}, v^{(i)}) \rightarrow x^{(i)}}$. For simplicity, we will omit the video index notation in the subsequent sections\footnote{In this paper, superscripts denote video indices, while subscripts usually refer to diffusion/flow timesteps.}.

In contrast to conventional approaches that perform the mapping through regression, our proposed framework leverages generative models.
The core component is our Separation U-Net $\boldsymbol{f_\theta}$ (detailed in \cref{subsec:network}), conditioned on the mixture magnitude $x^{mix}$ and the visual stream $v$.
In the DDPM variant (DAVIS), the framework employs the forward process (\cref{eq:ddpm_forward}) to noise the target spectrogram $x_0$ to $x_T$. As depicted in \cref{fig: framework}, the reverse process (\cref{eq:ddpm_reverse}) then uses the Separation U-Net, parameterized as $\epsilon_\theta$, to iteratively denoise a latent variable $x_T \sim \mathcal{N}(\mathbf{0},\mathbf{I})$ back to the separated magnitude $x_0$, predicting the noise $\epsilon$ at each step conditioned on $x^{mix}$ and $v^{(i)}$.
In the FM variant (DAVIS-Flow), we utilize the \textbf{same} Separation U-Net architecture $\boldsymbol{f_\theta}$, but train it as a conditional vector field approximator $v_\theta$. This network learns to map a sample $x_t$ on the probability path, evolving from noise $x_0$ to the target spectrogram $x_1$, to its velocity $v_\theta(x_t, t, \boldsymbol{c})$, conditioned on $t$, $x^{mix}$ and $v^{(i)}$ (forming $\boldsymbol{c}$). 

\subsection{Proposed Separation U-Net Architecture}
\label{subsec:network}
The Separation U-Net architecture described below serves as the core conditional generative model for both the DAVIS and DAVIS-Flow variants, with its design addressing key challenges in audio-visual separation.

Diffusion models often use U-Net-like~\citep{ronneberger2015u} architectures, which excel at capturing multi-level feature representations and maintaining the output shape identical to the input. These properties also make them well-suited for the audio-visual separation task in the network aspect. However, naively applying existing conditional diffusion models to audio-visual separation is ineffective, as they are typically designed for image-to-image translation~\citep{meng2021sdedit} or text-to-image synthesis~\citep{saharia2022photorealistic,rombach2022high}. These models utilize different condition mechanisms than those required for audio-visual tasks, and they are not tailored to address the unique characteristics of audio-visual data. Therefore, the development of a specialized audio-visual separation network for diffusion models is essential. 
In this context, we revisit the challenges that need to be addressed: (1) Similar frequency patterns commonly exist even in temporally distant time frames, which necessitates the network to capture both long-range dependencies across time and frequency dimensions, and thus pure convolution~\citep{zhao2018sound,gao2019co} may fall short. (2) Real-world videos often have mismatched visual and audio content. Extracting visual condition~\citep{tian2021cyclic,dong2023clipsep} without considering the possible unrelated audio-visual content can potentially lead to less discriminative visual cues. (3) Establishing precise audio-visual associations is crucial, but directly concatenating visual and audio embeddings at the bottleneck~\citep{gao2019co} lacks the ability to foster further interactions between the two modalities. 

To address these challenges, we propose a novel and specialized Separation U-Net in our generative framework that incorporates Convolution-Attention blocks to learn both local and global time-frequency associations, introduce a simple yet effective temporal transformer to aggregate the frame features and devise an audio-visual feature interaction module to enhance association learning by enabling interactions between audio and visual modalities.

\begin{figure*}[t]
\begin{center}
\includegraphics[width=1\linewidth]{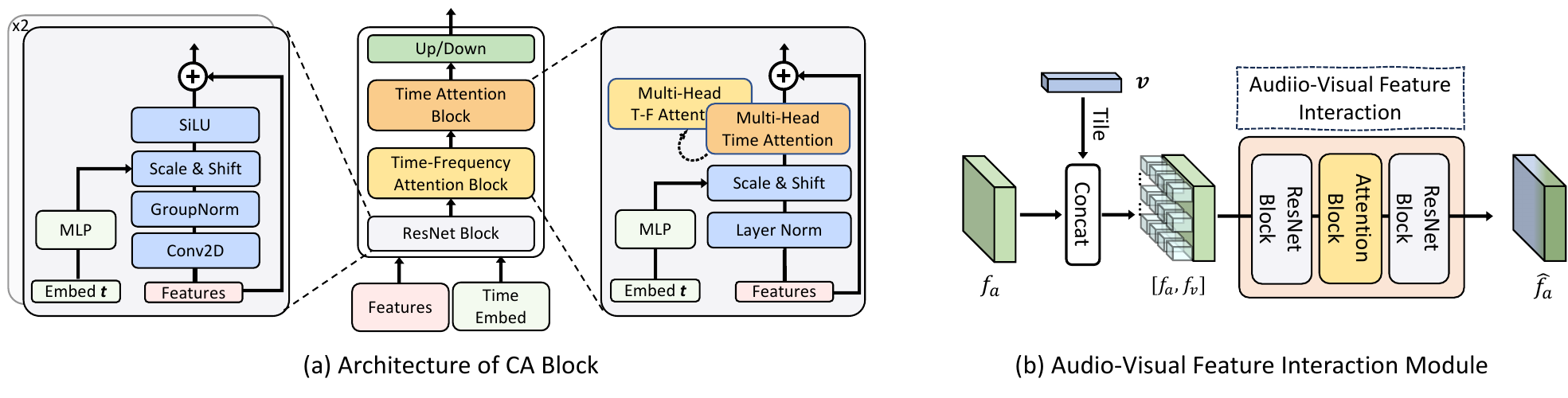}
\end{center}
  \caption{\textbf{Illustrations on (a) CA block: } 
  It operates by taking audio feature maps and a time embedding $\boldsymbol{t}$ as inputs. Each sub-block, except the up/down sampling layer, is conditioned on $\boldsymbol{t}$. 
    ResNet and attention blocks are stacked to capture local and non-local audio contexts;
    \textbf{(b) Audio-Visual Feature Interaction Module:} It functions by replicating and concatenating $\boldsymbol{v}$ with $\boldsymbol{f_a}$, and uses two identical ResNet blocks and an attention block to process the concatenated features.
    }

\label{fig: ca}
\end{figure*}

\noindent\textbf{Encoder/Decoder Designs.} Our proposed Separation U-Net architecture consists of an encoder and a decoder, linked by an audio-visual feature interaction module. Both the encoder and decoder comprise five CA blocks. Initially, we concatenate the noisy input $x_t$ with the mixture $x^{mix}$ along the channel dimension and use a 1x1 convolution to project it into the feature space (another 1x1 convolution to convert the decoder output back to magnitude). As depicted in \cref{fig: ca}(a), each CA block consists of a ResNet block, a Time-Frequency Attention block, and a Time Attention block.
Following this, a down-sample or an up-sample layer with a scale factor of 2 is used.
Concretely, we build the ResNet block using WeightStandardized 2D convolution~\citep{qian2020multiple} along with GroupNormalization~\citep{wu2018group} and SiLU activation~\citep{elfwing2018sigmoid}. To incorporate the time embedding $\boldsymbol{t}$ as a condition, a Multi-Layer Perceptron (MLP) is used to generate $\boldsymbol{t}$-dependent scaling and shifting vectors for feature-wise affine transformation~\citep{dumoulin2018feature}.
We also adopt an efficient form of attention mechanism~\citep{shen2021efficient} for implementing the Time-Frequency Attention block.
To enhance the long-range time dependency modeling, a Time Attention block is then appended. In practice, we follow the design in \citep{wang2023tf}, which includes Pre-Layer Normalization and Multi-Head Attention along the time dimension within the residual connection.
The down-sample and up-sample layers are simply 2D convolutions with a stride of 2. As a result, we can obtain audio feature maps $\boldsymbol{f_a} \in \mathbb{R}^{C \times \frac{T}{32} \times \frac{F}{32}}$ at the bottleneck, where $C$ represents the number of channels.

\noindent\textbf{Timestep Embedding.} In both DDPM and FM, the timestep embedding serves to inform the model about the current position $t$ within the generation process. As shown in \cref{fig: framework}, timestep $t$ is specified by the sinusoidal positional encoding~\citep{vaswani2017attention} and further transformed by an MLP, which will be passed to each CA block as a timestep condition. Note that for DDPM, $t \in \{1, ..., T\}$, while for FM, we typically normalize $t \in [0, 1]$.

\noindent\textbf{Visual Condition Aggregation.}
Not all frames in a video will be attributable to the synchronized audio. To account for unaligned visual content, we incorporate a shallow transformer to effectively aggregate the visual condition. Concretely, we extract frame features $\{\boldsymbol{I}_j\}_{j=1}^K$
from the visual stream $v$ using a pre-trained visual backbone $\mathbf{Enc_v}$, where $\boldsymbol{I}_j \in \mathbb{R}^{C}$. 
We apply a self-attention temporal transformer $\phi(\cdot)$ to aggregate raw visual frame features, resulting in $\{\boldsymbol{\hat{I}}_j\}_{j=1}^K = \phi(\{\boldsymbol{I}_j\}_{j=1}^K)$. For the transformer design, we empirically find that a shallow transformer with three encoder layers and one decoder layer works well. The global visual embedding $\boldsymbol{v}$ is then computed by averaging the temporal dimension of $\boldsymbol{v} = \frac{1}{K}\sum_{j=1}^K{\boldsymbol{\hat{I}}_j}$. 

\noindent\textbf{Audio-Visual Feature Interaction Module.} The key to audio-visual separation lies in effectively utilizing visual information to separate visually-indicated sound sources. Therefore, the interaction between audio and visual modalities at the feature level becomes crucial. Existing approaches often concatenate audio and visual features at the bottleneck~\citep{gao2019co,chatterjee2021visual} and pass them to the decoder for further fusion. This design, however, imposes a dual task on the decoder: to integrate visual cues while simultaneously reconstructing the audio signal. We hypothesize that enabling further audio-visual interaction at the bottleneck could potentially enhance the separation performance. To this end, we explore different interaction manners and propose an audio-visual feature interaction module to improve this capability (see \cref{tab:fim}). 
We spatially tile $\boldsymbol{v}$ to match the shape of $\boldsymbol{f_a}$, resulting in visual feature maps $\boldsymbol{f_v}$.  Subsequently, the audio and visual feature maps are concatenated along channel dimension and fed into the feature interaction module (FIM): $\hat{\boldsymbol{f_a}}:= \mathbf{FIM}([\boldsymbol{f_a}, \boldsymbol{f_v}])$, where $\hat{\boldsymbol{f_a}} \in \mathbb{R}^{C \times \frac{T}{32} \times \frac{F}{32}}$. The details of the module are illustrated in \cref{fig: ca}(b), including two ResNet blocks and a Time-Frequency Attention block to facilitate capturing audio-visual associations within both local and global regions.

\subsection{Training Pipeline}
\label{subsec:training}

Given the sampled audio-visual pairs from the dataset, we first adopt the ``mix and separate'' strategy and compute the magnitudes $x^{(1)}, x^{(2)}, x^{mix}$ with STFT. To align with the frequency decomposition of the human auditory system, we apply a logarithmic transformation to the magnitude spectrogram, converting it to a log-frequency scale. Additionally, we ensure consistent scaling by multiplying log-frequency magnitudes with a scale factor $\sigma$ and clipping the values to fall within the range $[0, 1]$. The visual frames are encoded to embeddings $\boldsymbol{v^{(1)}}$,$\boldsymbol{v^{(2)}}$. Let $\boldsymbol{c}^{(i)} = (x^{mix}, \boldsymbol{v}^{(i)})$ denote the conditional context for separating source $i$.

\noindent\textbf{DAVIS Training.}
For the DDPM variant, we train the Separation U-Net $\epsilon_\theta$ to predict the noise added during the forward diffusion process. Taking video (1) as an example, we sample $t$ uniformly from $\{1,..., T\}$, sample noise $\epsilon \sim \mathcal{N}(0, I)$, and compute the noised target $x^{(1)}_t$ using \cref{eq:ddpm_forward}. The network $\epsilon_\theta$ receives $x^{(1)}_t$, $t$, and the condition $\boldsymbol{c}^{(1)}$ as input. While the standard DDPM loss uses $\mathcal{L}_2$ (\cref{eq:ddpm_loss}), we hypothesize that the $\mathcal{L}_1$ loss is more robust to the skewed distribution of magnitude spectrograms, which often contain silent regions with near-zero values. Thus, we optimize the following objective:
\begin{equation}
    \mathcal{L}^{(1)}_{DDPM-L1}(\theta) = \mathbb{E}_{t, x^{(1)}_0, \epsilon}[ ||\epsilon - \epsilon_\theta(x^{(1)}_t, t, \boldsymbol{c}^{(1)})||_1].
\label{eq:ddpm_l1_loss}
\end{equation}
In practice, we use both videos (1) and (2) for training, and the final loss term is $\mathcal{L}_{DAVIS} =\mathcal{L}^{(1)}_{DDPM-L1}(\theta) +\mathcal{L}^{(2)}_{DDPM-L1}(\theta)$.

\vspace{1mm}
\noindent\textbf{DAVIS-Flow Training.}
For the FM variant, we take the same Separation U-Net, but train it as $v_\theta$ to predict the vector field of the conditional probability path. We use the linear probability path $x^{(1)}_t = t x^{(1)}_1 + (1-t) x_0$ for $t \in [0, 1]$, where $x^{(1)}_1$ is the target clean spectrogram for video (1) and $x_0 \sim \mathcal{N}(0, I)$ is Gaussian noise. The corresponding target vector field is $u_t(x^{(1)}_t|x^{(1)}_1) = x^{(1)}_1 - x_0$. 
The network $v_\theta$ receives $x^{(1)}_t$, $t$, and the condition $\boldsymbol{c}^{(1)}$ as input. Similar to the DDPM case, we opt for an $\mathcal{L}_1$ loss for robustness:
\begin{multline}
\mathcal{L}^{(1)}_{FM\text{-}L1}(\theta)
  = \mathbb{E}_{t, x_0, x^{(1)}_1}\Bigl[
      \|v_\theta(x^{(1)}_t, t, \boldsymbol{c}^{(1)})\\
      - (x^{(1)}_1 - x_0)\|_1
    \Bigr].
\label{eq:fm_l1_loss}
\end{multline}
Again, we use both videos for training, with the final loss being $\mathcal{L}_{DAVIS-Flow} = \mathcal{L}^{(1)}_{FM-L1}(\theta) + \mathcal{L}^{(2)}_{FM-L1}(\theta)$.

\subsection{Inference}
\label{subsec:inference}
During inference, the goal is to generate the separated spectrogram $x^{(i)}$ given the mixture $x^{mix}$ and the visual condition $\boldsymbol{v}^{(i)}$ (denoted jointly as $\boldsymbol{c}^{(i)}$). The process differs for DAVIS and DAVIS-Flow.

For DAVIS, inference starts with sampling a latent variable $x_T \sim \mathcal{N}(0, I)$ and iteratively applying the reverse transition $p_\theta(x_{t-1}|x_t)$ using the trained network $\epsilon_\theta$ for $t=T, ..., 1$. This process denoises the latent variable conditioned on $\boldsymbol{c}^{(i)}$ to produce the target sample $x_0^{(i)}$. In practice, we employ DDIM~\citep{song2020denoising} to accelerate the sampling process.
For DAVIS-Flow, the inference process involves solving the learned conditional probability flow ODE: ${dx_t}/{dt} = v_\theta(x_t, t, \boldsymbol{c}^{(i)})$. Starting from a noise sample $x_0 \sim \mathcal{N}(0, I)$, we integrate this ODE from $t=0$ to $t=1$ using the Euler solver. The result at $t=1$, denoted $x_1^{(i)}$, is the generated separated spectrogram. Usually, FM inference can potentially be faster than DDIM sampling as it may require fewer function evaluations (solver steps). 

\noindent\textbf{Silence Mask-Guided Sampling.}
As the goal of separation is to predict the individual sound from the mixture, an observation is drawn: silent time frames in the mixture should also be silent in the separated sound.
Therefore, we incorporate a guidance mechanism leveraging this observation to retain silent time frames in mixture $x^{mix}$ into the separated sound $x^{(i)}$. This prevents the model from ``hallucinating'' sound in silent regions and enforces consistency with the input mixture, akin to mask-based methods. During each step of the generation process, we apply the following refinement:

Let $x^{pred}_t$ be the direct prediction from the model at the current step $t$. We compute a corresponding noisy version of the mixture, $x^{mix}_{t}$, perturbed to roughly match the noise level of $x^{pred}_t$ (using either \cref{eq:ddpm_forward} or \cref{eq:fm_forward}). 
Given the perturbed mixture and the direction prediction, the denoising process can be updated as:
\begin{subequations}
    \begin{align}
        m &= [x^{mix} < \delta_{silence}], \label{eq:silence_mask} \\
        \Hat{x}^{pred}_{t} &= m \odot x^{mix}_{t} + (1 - m) \odot x^{pred}_t. \label{eq:silence_guidance}
    \end{align}
\end{subequations}  
Here, $m$ is the silence mask derived from the original clean mixture $x^{mix}$ using a threshold $\delta_{silence}$ (a hyperparameter and is set to 0.002). 
The refined prediction $\Hat{x}^{pred}_{t}$ replaces the silent regions identified by $m$ with the corresponding mixture content, while keeping the model's prediction elsewhere. This refined $\Hat{x}^{pred}_{t}$ is then used as the input for the next step in the iterative sampling step. This sampling strategy improves separation performance (see \cref{tab:silence}) by enforcing mixture consistency.
\Revision{This slience mask guidance is \emph{analogous} to discriminative
separation methods, which apply a time–frequency mask to the mixture spectrogram
so that silent regions are forced to zero. By explicitly injecting mixture
information into silent bins during generation, our formulation transfers this
advantage to diffusion-based separators.}

Finally, the waveform $s_i$ for the separated sound is reconstructed by applying the inverse STFT using the magnitude prediction and the phase from the original mixture $x^{mix}$. The training and inference pseudo codes can be found in \cref{alg:unified_training,alg:unified_inference}.
\section{Experiments}

\begin{figure*}[t]
    \centering
    \includegraphics[width=0.99\textwidth]{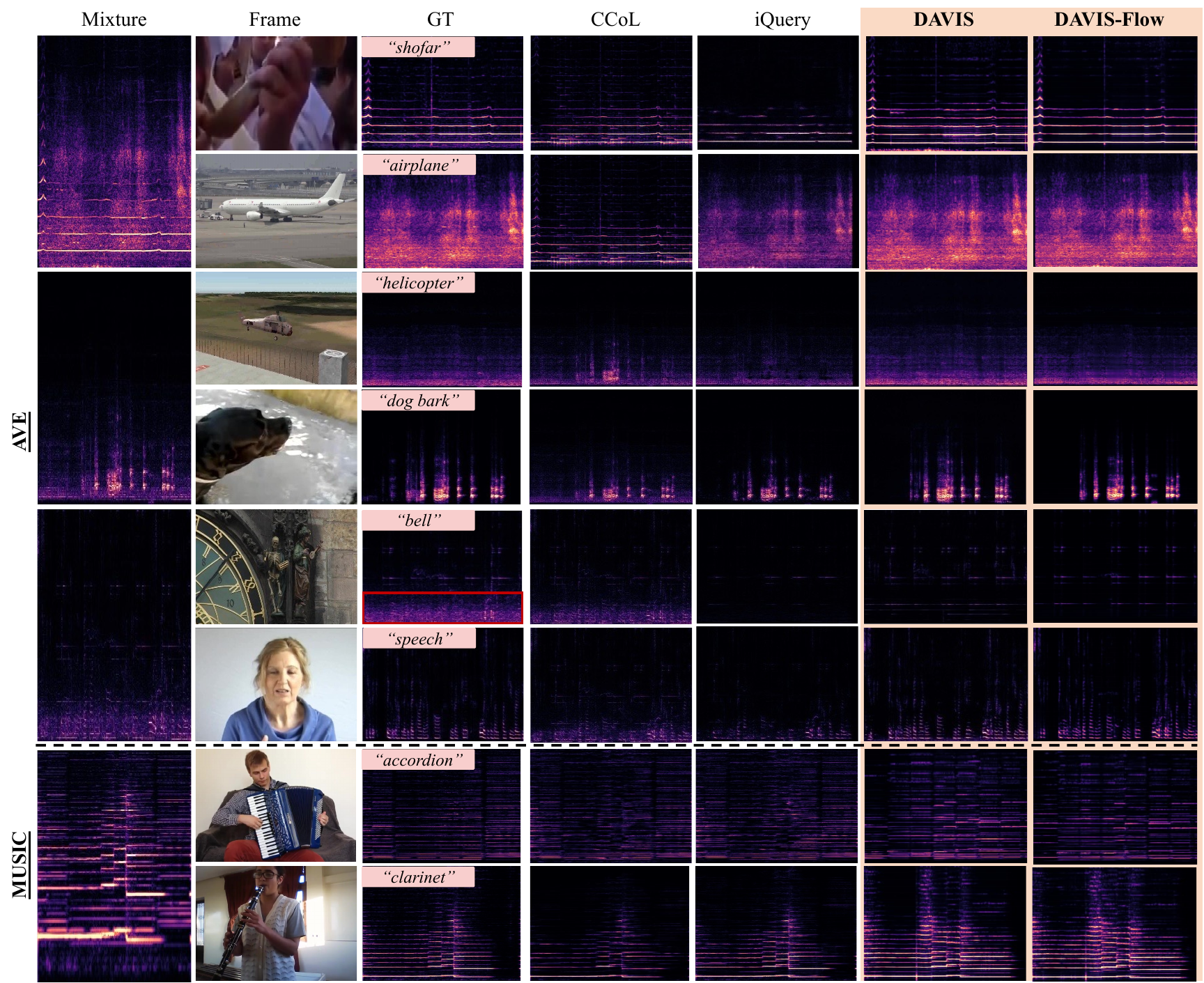}
    \caption{Visualizations of audio-visual separation results on the AVE (the top three mixtures) and MUSIC (the last mixture) datasets. Two sounds are mixed, and reference frames are provided to guide the separation. The comparison is shown between the predictions made by DAVIS (ours), DAVIS-Flow (ours) iQuery~\citep{chen2023iquery}, and CCoL~\citep{tian2021cyclic} with the ground truth. DAVIS and DAVIS-Flow can effectively separate sound mixtures from various categories, such as \textit{airplane}, \textit{rats}, and \textit{dog barking}.}
    \label{fig:vis}
\end{figure*}

\begin{table*}[!t]
    \centering
    \footnotesize
    \caption{Comparison of our method to other audio-visual separation approaches on the AVE and MUSIC test set. The top three results
    are highlighted in \textcolor{top1}{red}, \textcolor{top2}{orange}, and \textcolor{top3}{yellow}, respectively. The
    results noted by ${}^\dag$ are obtained from \citep{chen2023iquery,zhu2022visually}. 
Note that audio in AVE could include off-screen sounds and background noise, which may reduce the accuracy of the reported metrics.}
    \begin{tabularx}{0.95\textwidth}{lXrrrXrrr}
        \toprule
        & & \multicolumn{3}{c}{AVE~\citep{tian2018audio}}                                        & & \multicolumn{3}{c}{MUSIC~\citep{zhao2018sound}} \\
                            \cmidrule(lr){3-5}                                                        \cmidrule(lr){7-9}
        Methods    & & SDR $\uparrow$    & SIR $\uparrow$    & SAR $\uparrow$ & & SDR $\uparrow$    & SIR $\uparrow$    & SAR $\uparrow$\\
        \midrule
        NMF-MFCC${}^\dag$~\citep{spiertz2009source} & &- & -& -& & $0.92$ & $5.68$ & $6.84$ \\
        Sound-of-Pixels${}^\dag$~\citep{zhao2018sound} & & $1.21$ & $7.08$ & $6.84$ & & $4.23$ & $9.39$ & $9.85$ \\
        Co-Separation${}^\dag$~\citep{gao2019co} & &- & -& -& & $6.54$ & $11.37$ & $9.46$ \\
        Sound-of-Motions${}^\dag$~\citep{zhao2019sound} & & $1.48$ & $7.41$ & $7.39$  & &- & -& - \\
        Minus-Plus${}^\dag$~\citep{xu2019recursive} & & $1.96$ & $7.95$ & $8.08$ & &- & -& - \\
        Cascaded Filter${}^\dag$~\citep{zhu2020visually} & & $2.68$ & $8.18$ & $8.48$ & &- & -& - \\
        CCoL${}^\dag$~\citep{tian2021cyclic} & &- & -& -& & $7.74$ & $13.22$ & $11.54$ \\
        AMnet${}^\dag$~\citep{zhu2022visually} & & $3.71$ & \cellcolor{top2}$9.15$ & \cellcolor{top2}$11.00$ & &- & -& -\\
        iQuery${}^\dag$~\citep{chen2023iquery} & & \cellcolor{top2}$5.02$ & $8.21$ & \cellcolor{top1}$12.32$ & & \cellcolor{top3}$11.17$ & \cellcolor{top3}$15.84$ & \cellcolor{top3}$14.27$ \\
        \midrule
         \textbf{DAVIS (ours)} & & \cellcolor{top3}$4.86$  & \cellcolor{top3}$9.13$  & \cellcolor{top3}$9.92$ & & \cellcolor{top2}${11.61}$ & \cellcolor{top1}${18.36}$ & \cellcolor{top2}${14.70}$ \\
         \textbf{DAVIS-Flow (ours)} & & \cellcolor{top1}$5.66$ & \cellcolor{top1}$10.62$  & \cellcolor{top3}$10.63$ & & \cellcolor{top1}${12.01}$ & \cellcolor{top2}${18.25}$ & \cellcolor{top1}${15.46}$ \\
        \bottomrule
    \end{tabularx}
    \label{tab:ave_music}
\end{table*}

\subsection{Experimental Setup}

\noindent \textbf{Datasets.}
Our model demonstrates the ability to handle mixtures of diverse sound categories. To evaluate our approach, we use  AVE~\citep{tian2018audio} and MUSIC~\citep{zhao2018sound} datasets, which cover musical instruments and open-domain sounds. The evaluation settings are described in detail below:
   AVE~\citep{tian2018audio} contains 4143 10-second videos, including 28 diverse sound categories, such as \textit{Church Bell}, \textit{Barking}, and \textit{Frying}, among others. The AVE dataset presents greater challenges as the audio in these videos may not span the entire duration and can be noisy, including off-screen sounds (\textit{e.g.}, human speech) and background noise. 
    In addition to the AVE dataset, we also evaluate our proposed method on the widely-used MUSIC~\citep{zhao2018sound} dataset, which includes 11 musical instrument categories: accordion, acoustic guitar, cello, clarinet, erhu, flute, saxophone, trumpet, tuba, violin, and xylophone. All the videos are clean solo and the sounding instruments are visible. 
    For both datasets, we follow the same train/validation/test splits as in \citep{chen2023iquery,zhu2022visually}.

\noindent\textbf{Baselines.} To the best of our knowledge, we are the first to adopt a generative model for the audio-visual source separation task. Thus, we compare DAVIS against the following discriminative methods: \textit{NMF-MFCC}~\citep{spiertz2009source} which is an audio-only separation method; \textit{Sound of Pixels}~\citep{zhao2018sound} and \textit{Sound of Motions}~\citep{zhao2019sound} that learn ratio mask predictions with a 1-frame-based model or with motion as condition; \textit{Multisensory}~\citep{owens2018audio} that separates mixtures based on learning discriminative audio-visual representations; \textit{Minus-Plus}~\citep{xu2019recursive} that separates sounds by recursively eliminating high-energy components from the sound mixture; \textit{Cascaded Filter}~\citep{zhu2020visually} which separates sounds in a multi-stage manner; \textit{Co-Separation}~\citep{gao2019co} that takes a single visual object as the condition to perform mask regression; \textit{Cyclic Co-Learn} (CCoL)~\citep{tian2021cyclic} which jointly trains the model with sounding object visual grounding and visually-guided sound source separation tasks; \textit{AMnet}~\citep{zhu2022visually} which is a two-stage framework modeling both appearance and motion;
\textit{iQuery}~\citep{chen2023iquery} that adapts the maskformer architecture for audio-visual separation and achieves the current state-of-the-art (SOTA) results. 

\noindent\textbf{Evaluation Metrics.}
To quantitatively evaluate the audio-visual sound source separation performances, we use the standard metrics~\citep{zhao2018sound,tian2021cyclic,gao2019co}, namely: Signal-to-Distortion Ratio (SDR), Signal-to-Interference Ratio (SIR), and Signal-to-Artifact Ratio (SAR). We adopt the widely-used mir\_eval library~\citep{raffel2014mir_eval} to report the standard metrics. Note that SDR and SIR evaluate the accuracy of source separation, whereas SAR specifically measures the absence of artifacts~\citep{gao2019co}.

\subsection{Comparisons with State-of-the-art}
\label{subsec:comparison}

To evaluate the effectiveness of our proposed generative framework, we conduct a comparative study pitting both DAVIS and DAVIS-Flow against existing state-of-the-art audio-visual separation methods on the AVE and MUSIC datasets. The quantitative results of this comparison are summarized in \cref{tab:ave_music}. Overall, our findings underscore the advantages offered by employing a generative modeling approach for this task
Comparing our methods to the strong baseline iQuery~\cite{chen2023iquery}, DAVIS achieves comparable performance in terms of SDR on average, while the DAVIS-Flow variant demonstrates even stronger results, surpassing iQuery across nearly all evaluation metrics. It is important to note that the results for the original DAVIS on the AVE dataset are less definitive compared to those on the MUSIC dataset. We attribute this difference primarily to two factors: Firstly, iQuery utilizes ground truth class labels during both training and inference, providing a more powerful conditional signal than the video frames alone used by DAVIS. Secondly, the audio tracks in the original AVE dataset often contain significant off-screen sounds and background noise. Standard objective metrics such as SDR, SIR, and SAR may not fully or accurately reflect separation quality in the presence of such pervasive noise, even when the separation process successfully removes the target sound along with some background elements. Despite these inherent challenges of the AVE dataset and the conditional advantage of iQuery, DAVIS-Flow, our Flow Matching variant, consistently outperforms iQuery on both datasets evaluated and even surpasses the performance of the vanilla DAVIS model.
Qualitative visualizations presented in \cref{fig:vis} further support these quantitative findings, reinforcing the advantages of both DAVIS and DAVIS-Flow in producing high-quality separations. On the cleaner MUSIC dataset, both DAVIS and DAVIS-Flow consistently exhibit superior performance across various standard evaluation metrics, clearly outperforming the next best method, iQuery. These results collectively demonstrate the versatility of our DAVIS framework across datasets with differing characteristics and sound categories, a conclusion also supported by the subjective listening test discussed in \cref{subsec:qual_analysis}.

\begin{table}[t]
    \centering
    \caption{Ablation on CA block design. R, TF, and T denote ResNet, Time-Frequency, and Time Attention blocks, respectively. We highlight the setting used in this paper in \colorbox{mygray}{gray}.}
    \begin{tabular}{lccc|c}
        \toprule
        Block & SDR$\uparrow$ & SIR$\uparrow$ & SAR$\uparrow$  & \# Params (M)\\
        \midrule
        \{R, R, R\} & $9.03$ & $14.05$ &  $13.20$ & $51.76$ \\ 
        \{R, R, T\} & $11.78$ & $17.91$ & $15.44$ & $43.85$ \\ 
        \{R, R, TF\} & $11.50$ & $18.01$  & $15.21$ & $42.95$ \\   
        \rowcolor{mygray} \{R, TF, T\} & $11.88$ & $17.52$  & $16.12$ & $35.04$ \\   
        \bottomrule
    \end{tabular}
    \label{tab:ca}
    
\end{table}

\subsection{Ablation Studies}
\label{subsec:analysis}
We conduct ablations on the MUSIC validation set (unless specified) to examine the different components of DAVIS framework. 
\\
\textbf{Block Design.} We validate the effectiveness of our proposed CA block (shown in \cref{fig: ca}) by designing the following baselines: (a) using three consecutive ResNet blocks within the CA block, which only captures local time-frequency patterns; (b) replacing the last ResNet block with a Time Attention block; (c) replacing the last ResNet block with a Time-Frequency Attention block; and (d) replacing the last two ResNet blocks with Time-Frequency and Time attention blocks to enhance the capability of modeling long-range dependency.
The results in \cref{tab:ca} underscore the importance of learning both local and global contexts across time and frequency dimensions. Furthermore, the comparison of model sizes confirms that the improvements are not attributable to increased network capacity. 
\begin{table}[h]
    \centering
    \vspace{-4mm}
    \caption{Ablation study on Feature Interaction Module. We explore different ways of integrating audio and visual features.}
    \begin{tabular}{lccc|c}
        \toprule
        Fusion & & SDR$\uparrow$ & SIR$\uparrow$ & SAR$\uparrow$   \\
        \midrule
        Concat & & $10.85$ &  $17.62$ & $15.52$  \\  
        FIM (Point-wise) & & $11.06$  & $17.37$  & $15.44$  \\ 
        FIM (Local)        & & $11.56$  & $17.02$  & $16.28$  \\
        FIM (Global)       & & $11.23$  & $17.56$  & $15.84$  \\
        \rowcolor{mygray} FIM (Local{\&}Global) & & $11.88$  & $17.52$  & $16.12$  \\ 
        \bottomrule
    \end{tabular}
    \label{tab:fim}
    \vspace{-4mm}
\end{table}

\noindent\textbf{Audio-Visual Feature Interaction.} To validate the importance of effective audio-visual association learning for this task, we conduct an ablation study on the Feature Interaction Module. Specifically, we explore different ways of feature interaction: (a) direct concatenation of visual and audio features, (b) a three-layer MLP for point-wise fusion, (c) three ResNet blocks, (d) three attention blocks, and (e) a combination of ResNet and attention blocks. The results presented in \cref{tab:fim} show that naive concatenation of audio and visual features performs significantly poorly while enabling further interaction between them improves the results. Among all the designs, our proposed module achieves the best results by considering both local and non-local contexts.

\noindent\textbf{Effects of Silence Mask-Guided Sampling.} As shown in \cref{tab:silence}, we experiment with different thresholds for the silence mask-guided sampling method, which determines the proportion of re-used information from the mixture. While a high threshold may introduce leakage from non-silent regions (\textit{e.g.}, from the second sound), we show that carefully selecting the threshold value can boost the separation performance in the post-training stage compared to the baseline.

\begin{table}[!t]
    \centering
    \caption{The effect of silence mask-guided sampling strategy on the MUSIC test set.}
        \label{tab:silence}
    \begin{tabularx}{\linewidth}{lXccc}
        \toprule
        $\delta_{silence}$ & & SDR$\uparrow$ & SIR$\uparrow$ & SAR$\uparrow$   \\
        \midrule
        $0$ (baseline) & & $11.53$ &  $18.30$ & $14.74$  \\  
        \midrule
        $0.01$         & & $11.15$ &  $18.21$ & $14.69$  \\ 
        $0.001$        & & $11.50$ &  $18.26$ & $14.74$ \\
        \rowcolor{mygray} $0.002$ & & $11.61$ & $18.36$ & $14.70$  \\ 
        \bottomrule
    \end{tabularx}
    \vspace{-4mm}
\end{table}

\begin{figure}[b]
        \centering
        \includegraphics[width=0.99\columnwidth]{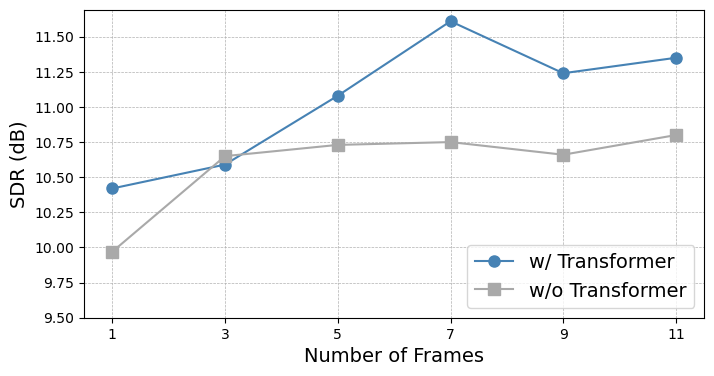}
        \captionof{figure}{Ablation on varying the number of frames to validate the effect of our proposed temporal transformer. }
        \label{fig:transformer}
\end{figure}

\noindent\textbf{Aggregated Visual Condition.} To study the impact of the visual condition, we vary the number of sampled frames and compare models with and without the temporal transformer, as shown in \cref{fig:transformer}. The results reveal that increasing the number of frames achieves a more informative visual condition and boosts separation quality. However, as the number of frames increases, noisy information may be introduced, leading to a decline in performance. We show that adopting a temporal transformer effectively alleviates this issue, resulting in better separation performance.

\begin{table}[!t]
  \centering
  \caption{Ablation for the choice of loss function on the MUSIC~\citep{zhao2018sound} and AVE~\citep{tian2018audio} datasets (test split).}
  \label{tab:loss_ave}
  \begin{tabularx}{\linewidth}{
      l 
      *{3}{>{\centering\arraybackslash}X}
      *{3}{>{\centering\arraybackslash}X}
    }
    \toprule
    Loss fn.\ & \multicolumn{3}{c}{MUSIC} & \multicolumn{3}{c}{AVE} \\
    \cmidrule(lr){2-4} \cmidrule(lr){5-7}
     & SDR$\uparrow$ & SIR$\uparrow$ & SAR$\uparrow$ 
     & SDR$\uparrow$ & SIR$\uparrow$ & SAR$\uparrow$ \\
    \midrule
    $\mathcal{L}_2$ 
      & 10.84 & 17.52 & 14.55 
      & 4.53  & 8.26  & 10.14 \\
    $\mathcal{L}_1$ 
      & \textbf{11.61} & \textbf{18.36} & \textbf{14.77} 
      & \textbf{4.86}  & \textbf{9.13}  & \textbf{9.92} \\
    \bottomrule
  \end{tabularx}
\end{table}
\noindent\textbf{$\mathcal{L}_1$ Training Loss}
\cref{tab:loss_ave} shows that an $\mathcal{L}_1$ loss in training the diffusion model performs better than an $\mathcal{L}_2$ loss for audio-visual separation on both MUSIC and AVE datasets. It's because of the presence of silent time frames in magnitude spectrograms, where the values are almost zero. This skewed data distribution renders the conventional $\mathcal{L}_2$ loss in diffusion models susceptible to error.

\subsection{More Analysis}
\label{subsec:qual_analysis}

\begin{figure}[!b]
    \centering
    \includegraphics[width=0.99\columnwidth]{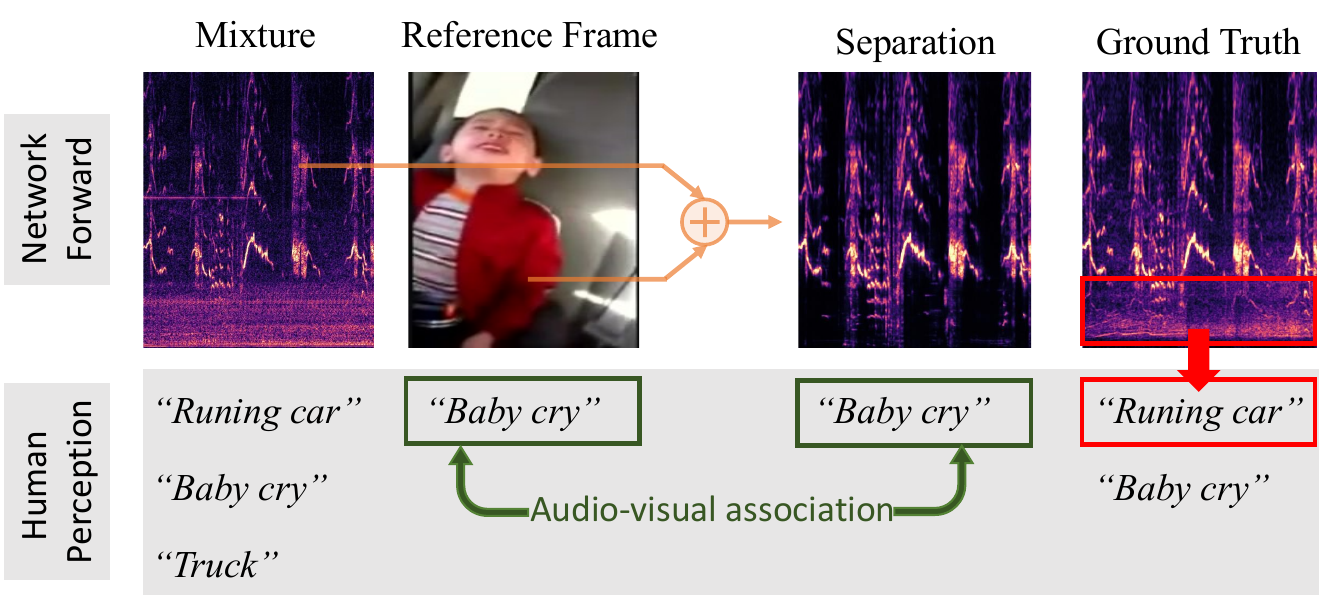}
    \caption{A visualization example showing that our DAVIS model can capture accurate audio-visual association.}
    \label{fig:association}
\end{figure}

\begin{figure}[!ht]
    \centering
    \includegraphics[width=0.25\textwidth]{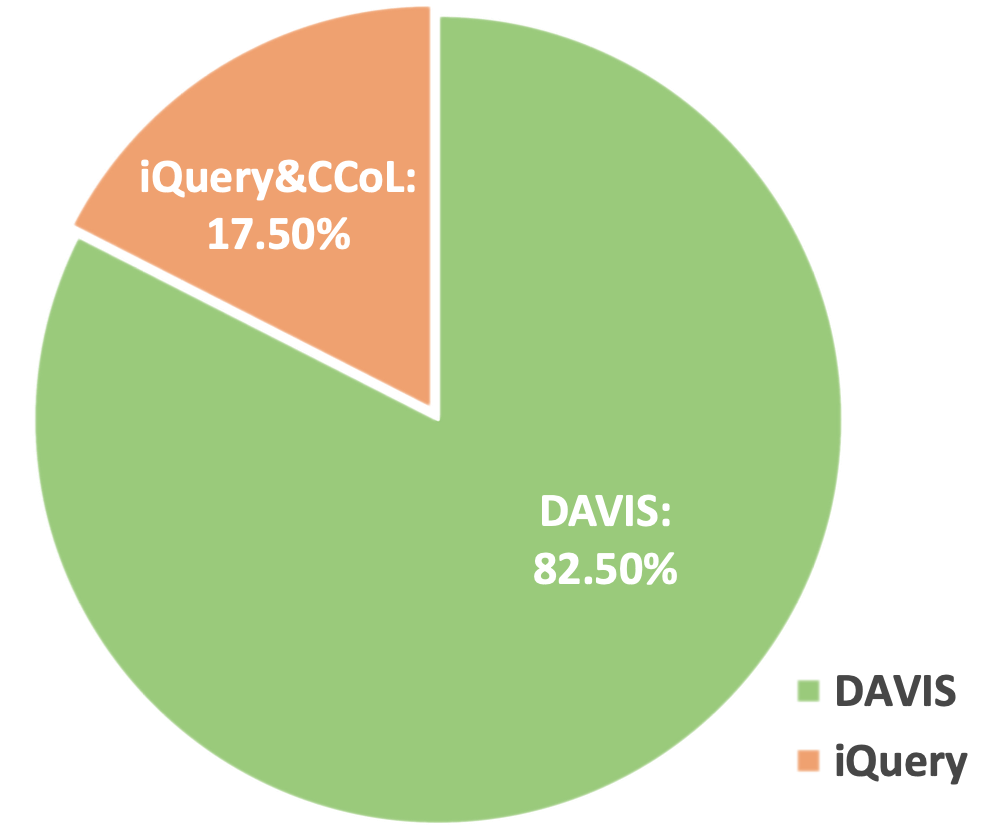}
    \caption{Human evaluation results for sound source separation on mixtures of different categories.}
    \label{fig:subjective}
\end{figure}

\begin{figure}[!b]
    \centering
    \includegraphics[width=0.49\textwidth]{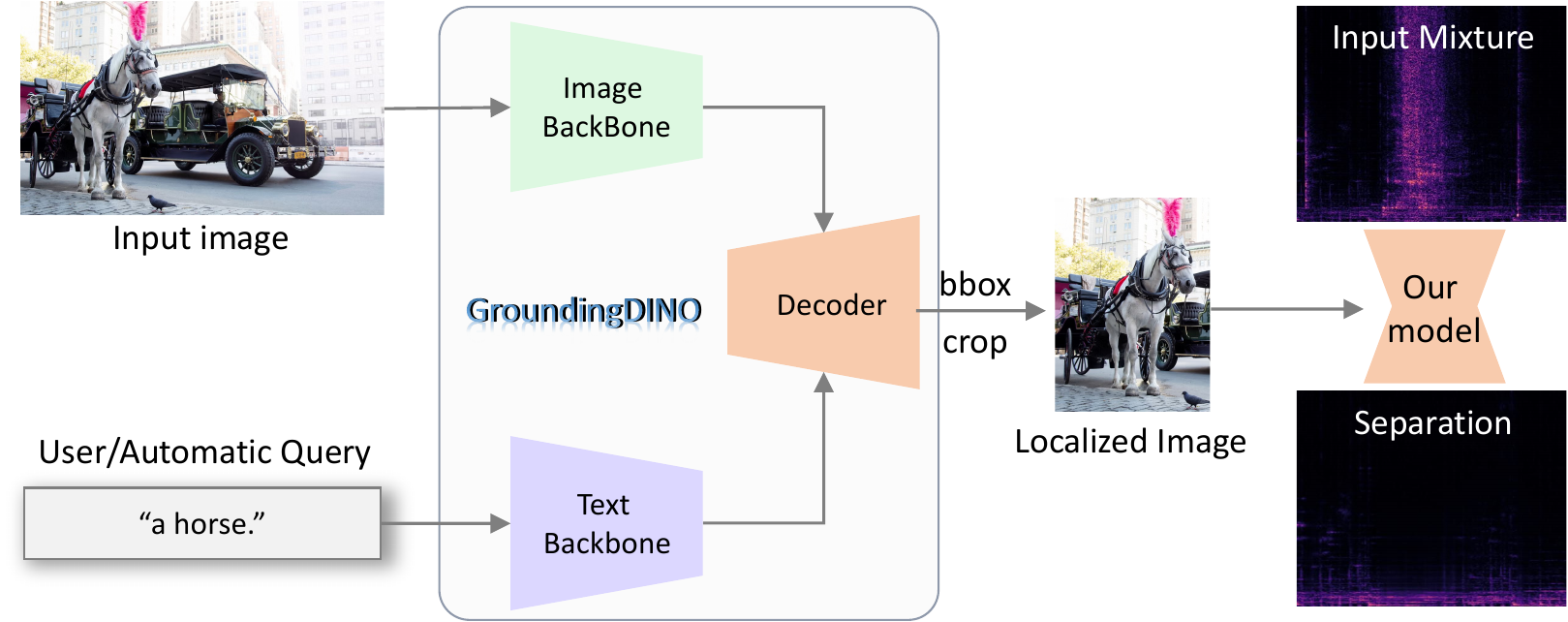}
    \caption{Pipeline for localized conditioning using GroundingDINO.  
    Given a text description and image frame, GroundingDINO detects the target object, enabling us to crop and encode it with CLIP for targeted audio–visual separation.}
    \label{fig:grounding}
\end{figure}

\noindent\textbf{Learned Audio-Visual Association.} 
To showcase the accuracy of our model's learned audio-visual associations, we mixed a ``Baby crying'' video clip with a ``Truck'' video clip from the AVE dataset.
As shown in \cref{fig:association}, the original baby video, as perceived by human listeners, also contains a running car sound, thus establishing a complicated audio-visual relationship. Our model successfully extracts the baby's crying sound while eliminating all irrelevant sounds, demonstrating DAVIS's ability to learn accurate audio-visual associations.

\begin{figure*}[!ht]
    \centering
    \includegraphics[width=0.85\textwidth]{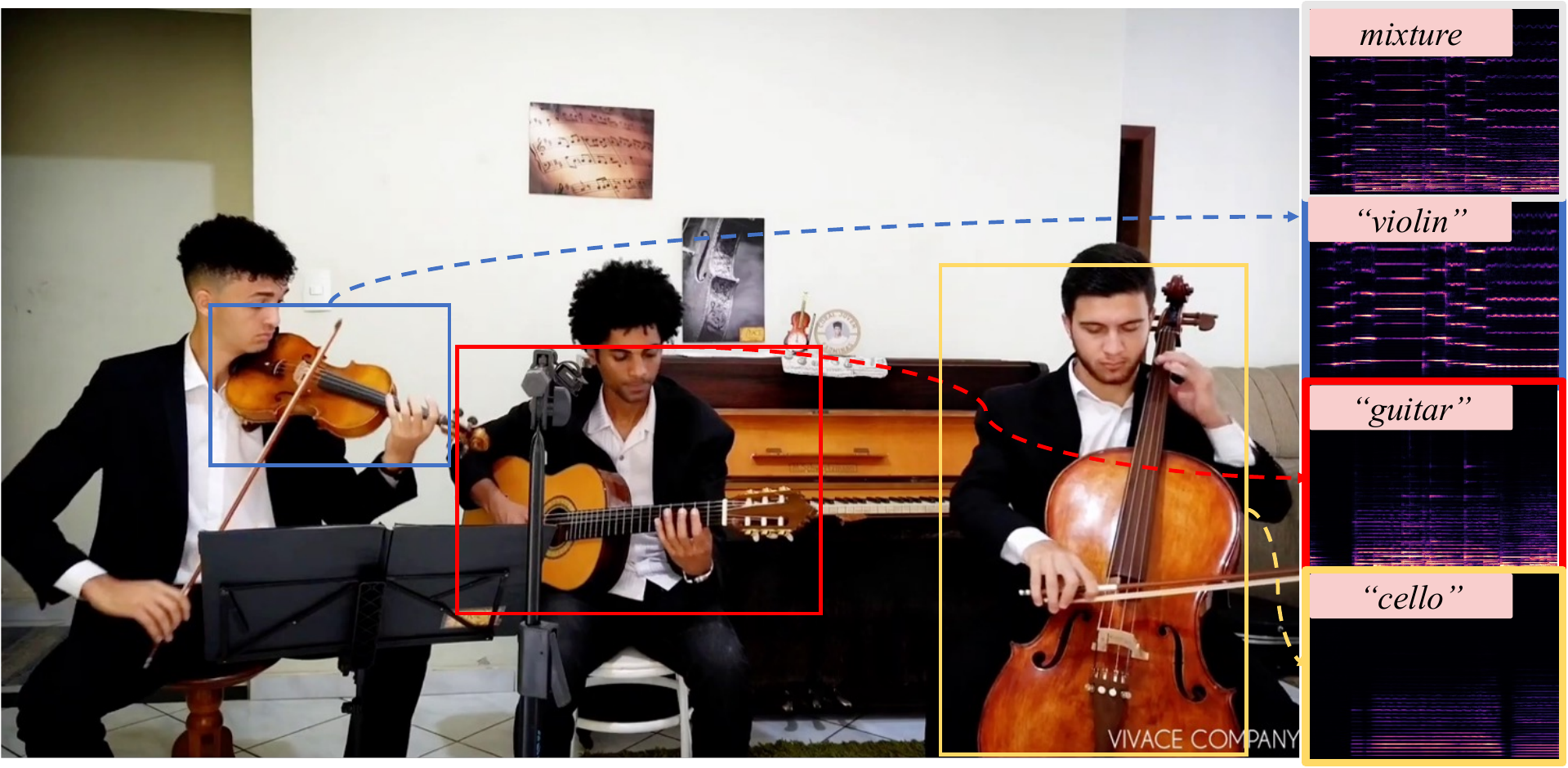}
    \caption{\textbf{Separation results on a real-world challenging three-source example}. 
    YouTube ID: \textit{R1DCTNEMibw}.
    }
    \label{fig:natural}
\end{figure*}

\noindent\textbf{Subjective Test.} We conduct a subjective test of separation results of our model and strong baselines iQuery/CCoL. 11 participants are asked to answer \textit{``Which separation result is closer to the ground truth audio and better matches the frame content?''}, with GT sound and frame as a reference. DAVIS outperformed the other baselines with a \textbf{winning rate of 82.5\%} from 176 results, as shown in the figure to the right.

\Revision{
\noindent\textbf{Localized CLIP Embeddings.}
In complex visual scenes, multiple objects may correspond to different sound sources.  
To separate the sound of a specific object, the model should leverage \emph{localized} visual cues rather than only global scene information.  
Although our main pipeline uses global CLIP visual embeddings for both training and inference, it can be readily extended to support localized conditioning.

Concretely, during inference we incorporate object-level grounding using GroundingDINO~\citep{liu2024grounding}.  
We assume that in a multi-object video, the target object is known, either specified directly by the user or identified automatically via an object detection or captioning model.  
Given the image frame and a textual description of the target object, GroundingDINO returns its bounding box.  
We then crop the image to this bounding box and feed the localized region into the CLIP image encoder to obtain a targeted embedding.  
This localized embedding replaces the global embedding as the condition for our separation model.  
The overall process is illustrated in \cref{fig:grounding}.

\begin{table}[!t]
  \centering
  \caption{Separation results of {DAVIS-Flow} with and without object-level grounding (GroundingDINO: box=0.7, text=0.3). 
  Train/Test columns indicate whether grounding is used (\cmark) or not (\xmark).}
  \label{tab:grounding}
  \setlength{\tabcolsep}{4pt}
  \begin{tabular*}{\linewidth}{@{\extracolsep{\fill}}lccccc}
    \toprule
    \textbf{Dataset} & \textbf{Train G.} & \textbf{Test G.} & \textbf{SDR}$\uparrow$ & \textbf{SIR}$\uparrow$ & \textbf{SAR}$\uparrow$ \\
    \midrule
    \multirow{2}{*}{MUSIC}
      & \xmark & \xmark & 12.01 & 18.25 & 15.46 \\
      & \xmark & \cmark & 11.75 & 18.16 & 15.29 \\
    \midrule
    \multirow{2}{*}{AVE}
      & \xmark & \xmark & 5.66  & 10.62 & 10.63 \\
      & \xmark & \cmark & 5.29  & 10.04 & 10.68 \\
    \bottomrule
  \end{tabular*}
\end{table}

\Cref{tab:grounding} reports results on the MUSIC and AVE datasets.  
Even though our model was trained exclusively with global CLIP embeddings, replacing them at inference with localized embeddings from GroundingDINO leads to only marginal performance changes.  
This demonstrates that our approach is robust to grounding and applicable to complex multi-object visual scenes.
}

\noindent\textbf{Qualitative Visualization on Natural Sound Mixture.}
To further demonstrate our model's effectiveness, we present a challenging real-world example of testing it on a natural sound mixture with multiple sounds (see Fig.~\ref{fig:natural}).

\begin{table}[!t]
    \centering
    \caption{Intra-class (same-category) separation on the MUSIC-based 100-mixture benchmark.}
    \label{tab:intra_class}
    \begin{tabularx}{\linewidth}{lXcc}
        \toprule
        \multicolumn{4}{c}{\textbf{MUSIC — Same-Class Mixtures}} \\
        \midrule
        \textbf{Method} & & \textbf{SDR}$\uparrow$ & \textbf{SIR}$\uparrow$  \\
        \midrule
        Mixture    & & 0.22 & 0.22  \\ 
        DAVIS-Flow & & \textbf{0.54} & \textbf{3.26}  \\  
        iQuery     & & -0.96 & 0.24 \\
        \bottomrule
    \end{tabularx}
\end{table}

\Revision{
\noindent\textbf{Intra-Class Separation.}
To probe the limits of our approach, we evaluate on challenging ``same-class'' mixtures where two instances of the \emph{same} category are combined (\textit{e.g.}, two violins). We construct a test set from MUSIC with 100 mixtures, each formed by mixing two clips from the same class. This setting is intrinsically difficult because the conditional signal (visual embeddings) is less discriminative across sources, yet it offers a fine-grained stress test.
We compare \mbox{DAVIS-Flow} against the unprocessed mixture and the strongest discriminative baseline, iQuery~\citep{chen2023iquery}. As shown in \cref{tab:intra_class}, \mbox{DAVIS-Flow} achieves non-trivial separation -- improving over the mixture -- whereas iQuery underperforms the mixture in SDR, indicating difficulty handling such intra-class ambiguity.}

\begin{figure}[!t]
 \centering
 \includegraphics[width=0.9\linewidth]{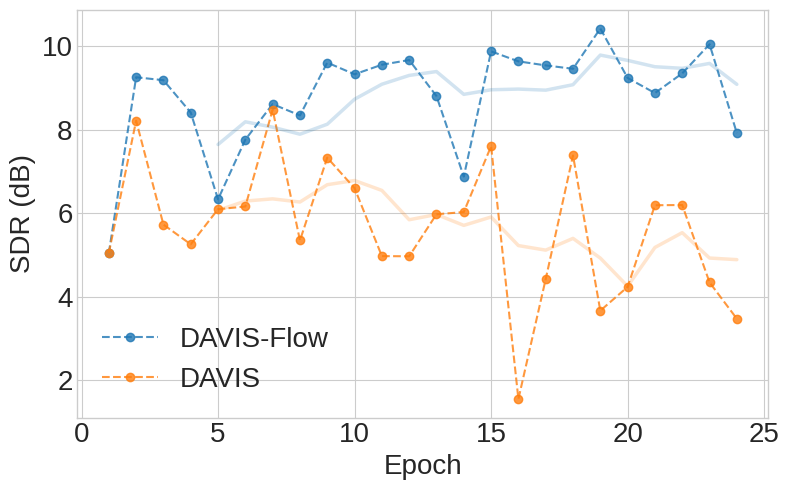}
 \caption{Early stage training dynamics comparison between vanilla DDPM-based DAVIS and DAVIS-Flow. Solid lines represent the smoothed training curves, while dashed lines show the per-epoch performance.} %
 \label{fig:training}
\end{figure}

\subsection{DAVIS vs. DAVIS-Flow: Efficiency Analysis}
\label{subsec:versu}

In this section, we compare the two distinct variants of our DAVIS framework to demonstrate that DAVIS-Flow, which leverages the Flow Matching paradigm, offers notable advantages in terms of both training speed and inference efficiency compared to the original DDPM-based DAVIS, thereby reducing the computational overhead.

\noindent\textbf{Training Dynamics.}
The training dynamics of vanilla DAVIS and DAVIS-Flow are compared in \cref{fig:training}. While both methods exhibit similar performance levels at the initial stages of training (specifically, after the first epoch), DAVIS-Flow consistently demonstrates faster performance improvement and more stable convergence behavior, characterized by less fluctuation in training metrics. Although the performance of DAVIS may occasionally reach that of DAVIS-Flow at certain points during training, the smoothed training curves presented clearly illustrate that DAVIS-Flow achieves more efficient convergence with enhanced stability over time.

\begin{figure}[!t]
 \centering
 \includegraphics[width=0.9\linewidth]{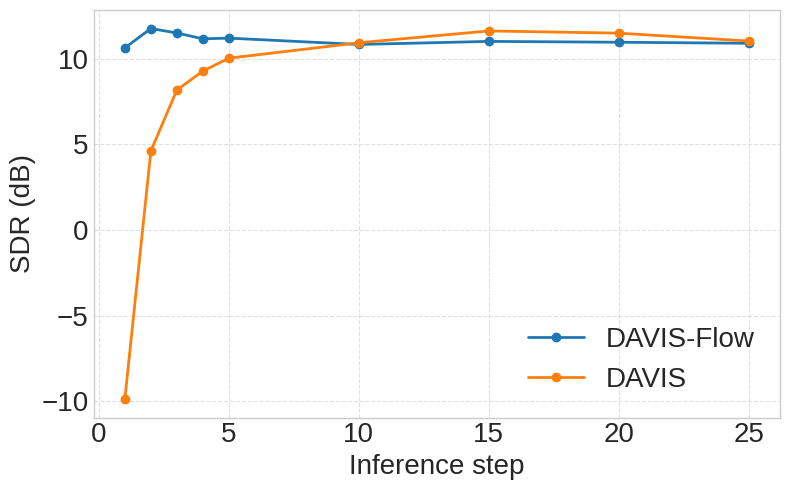}
 \caption{Ablation study on the number of sampling steps for DAVIS (with DDIM acceleration) and DAVIS-Flow (with Euler solver). Performance metric vs. number of steps $T \in \{1,2,3,4,5,10,15,20,25\}$.} %
\label{fig:step}
\end{figure}

\begin{figure*}[!ht]
    \centering
    \includegraphics[width=0.95\textwidth]{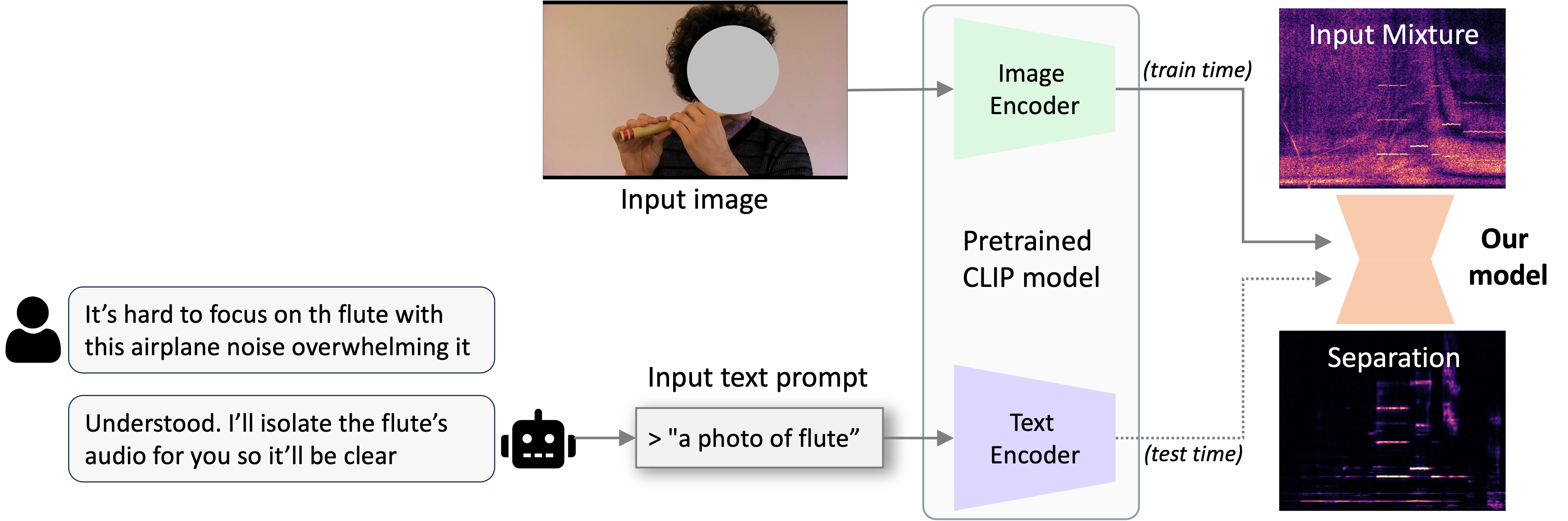}
    \caption{\textbf{Application zero-shot text-guided source separation.} During testing, text prompts are used as conditions for the DAVIS model, which is trained with images as conditions.}
    \label{fig:zeroshot}
\end{figure*}

\noindent\textbf{Inference Efficiency.}
We analyze the impact of the number of sampling steps on the performance of both DAVIS and DAVIS-Flow to assess their inference efficiency. \cref{fig:step} presents the results of this ablation study, showing the separation performance metric as a function of the number of steps, evaluated across $T \in \{1,2,3,4,5,10,15,20,25\}$. For this comparison, we employ DDIM~\citep{song2020denoising} to accelerate the sampling process in DAVIS and use the standard Euler solver for DAVIS-Flow. The results indicate that for the DDPM‐based DAVIS, utilizing a small number of inference steps ($T < 10$) is insufficient for achieving effective sound separation, with performance only converging and stabilizing at a higher number of steps ($T > 15$). In stark contrast, DAVIS-Flow reaches its peak performance with a significantly reduced number of steps ($T=2$), and further increasing the number of steps beyond this point does not lead to substantial performance gains. This analysis clearly highlights that while DAVIS can achieve satisfactory separation results, it necessitates a considerably larger number of inference steps (approximately $T=15$) compared to the highly efficient DAVIS-Flow, which performs optimally with just $T=2$ steps.

\Revision{
\noindent\textbf{Computational Cost Analysis.}
While DAVIS-Flow significantly reduces the required number of inference steps to just two, thereby lowering overall computational overhead, it remains more computationally intensive than discriminative models, which typically complete inference in a single step.

To quantify the computational cost, we compare the inference latency of our methods with that of the strong discriminative baseline iQuery~\citep{chen2023iquery} on an RTX 4090 GPU. As shown in \cref{tab:latency}, iQuery requires only one inference step and achieves a latency of 0.018 seconds, while DAVIS-Flow achieves low-latency inference at 0.156 seconds using two steps. DAVIS, on the other hand, requires more steps and thus incurs higher latency (\textit{e.g.}, 1.171 seconds at $T=15$ steps). Despite this, both DAVIS and DAVIS-Flow maintain model sizes comparable to iQuery.
}

\begin{table}[!t]
  \centering
  \caption{Inference latency and model size comparison between our methods and the strongest discriminative baseline iQuery~\cite{chen2023iquery}.}
  \label{tab:latency}
  \begin{tabularx}{\linewidth}{lccc}
    \toprule
    \textbf{Method} & \textbf{Params. (M)} & \textbf{Steps} & \textbf{Latency (s)} \\
    \midrule
    iQuery & 35.25 & 1 & 0.018 \\
    DAVIS & 35.04 & 15 & 1.171 \\
    DAVIS-Flow & 35.04 & 2 & 0.156 \\
    \bottomrule
  \end{tabularx}
\end{table}

\subsection{Application: Zero-Shot Text-guided Separation}
\label{subsec:zeroshot}
\begin{figure*}[!t]
    \centering
    \includegraphics[width=0.8\textwidth]{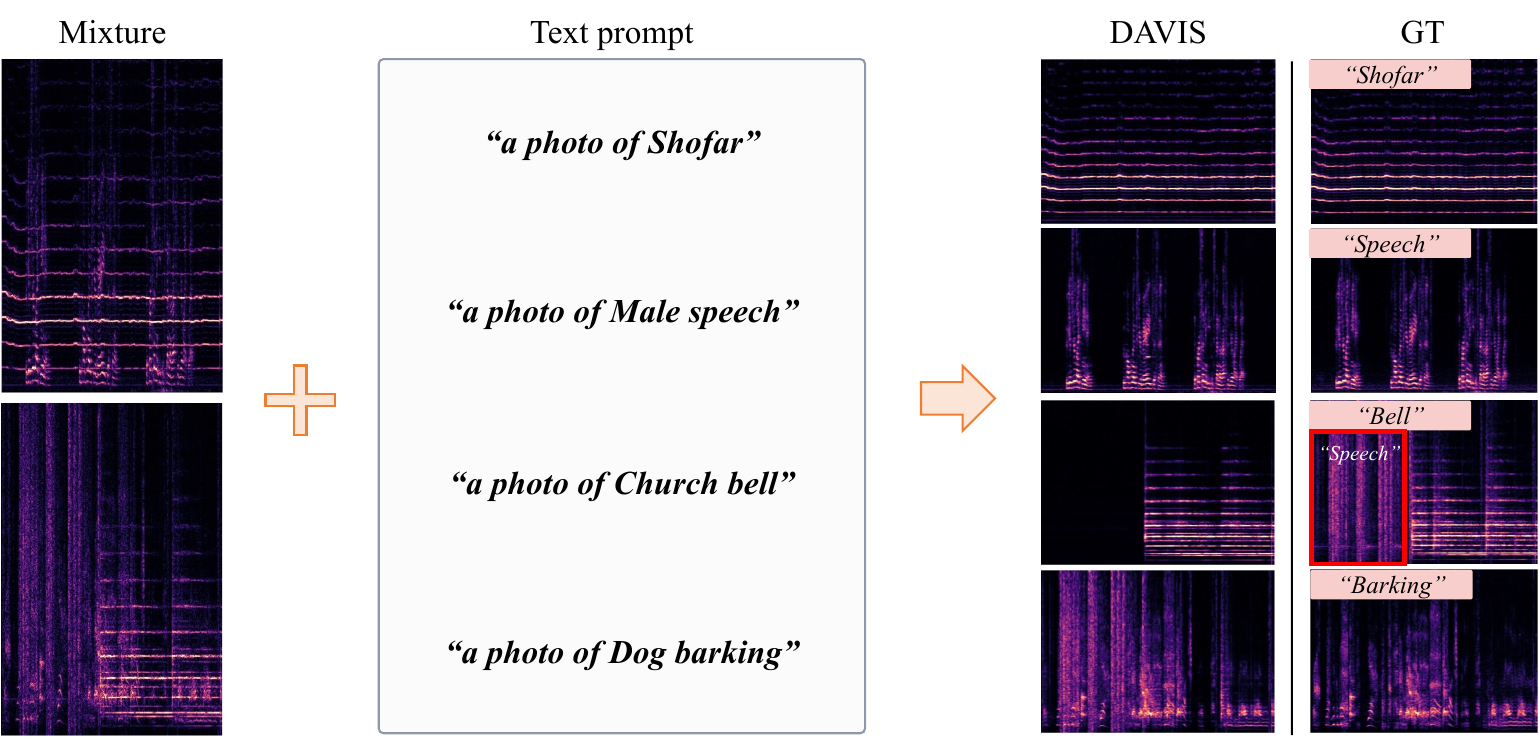}
    \caption{\textbf{Qualitative examples of zero-shot text-guided source separation.} Notably, in \textit{the third row} example, we observe the model's ability to capture precise audio-text correspondence by successfully filtering out the \textit{``speech''} sound. }
    \label{fig:textprompt}
\end{figure*}

Our model trained to capture the conditional distribution $p(x|\mathbf{v})$ can be employed for zero-shot inference from $p(x|\mathbf{t})$ where $\mathbf{t}$ represents the text description corresponding to the image $\mathbf{v}$. We achieve this by leveraging the well-established shared image-text embedding space from the CLIP~\citep{radford2021learning} model.
During the training phase, we extract the visual embeddings from the CLIP-image encoder used as conditions. Capitalizing on CLIP's robust ability to map semantically related images and text, we can substitute the visual features with text features generated by the CLIP-text encoder from a given text query during inference. This allows for effective zero-shot transfer. Furthermore, by integrating the parsing abilities of Large Language Models, our framework can be extended to enable user-instructed sound separation, allowing users to guide the separation process by providing natural language text prompts, as shown in \cref{fig:zeroshot}. 
We qualitatively evaluate the results of utilizing replacement conditioning for separation and find it to be surprisingly effective, as shown in \cref{fig:textprompt}. \Revision{We also present quantitative results in \cref{tab:text_combined}. Notably, even without explicit text–audio training, our zero-shot text-guided separation achieves performance comparable to the second-best discriminative methods -- AMnet on AVE and CCoL on MUSIC -- demonstrating the strong generalization and practicality of our approach.}

\begin{table}[!t]
    \centering
    \caption{Zero-shot text-guided separation performance on MUSIC and AVE test sets. 
    DAVIS-Flow (visual) uses visual embeddings as conditions, 
    while DAVIS-Flow (text) uses CLIP text features for zero-shot inference.}
    \label{tab:text_combined}
    \begin{tabularx}{\linewidth}{lXccc}
        \toprule
        \multicolumn{5}{c}{\textbf{MUSIC Test Set}} \\
        \midrule
        Method & &SDR$\uparrow$ &  SIR$\uparrow$ & SAR$\uparrow$ \\
        \midrule
        DAVIS-Flow (visual) & & {12.01} & {18.25} & {15.46} \\  
        DAVIS-Flow (text)   & &7.51 & 14.06 & 13.58 \\ 
        CCoL                & &7.74 & 13.22 & 11.54 \\
        \midrule
        \multicolumn{5}{c}{\textbf{AVE Test Set}} \\
        \midrule
        Method & & SDR$\uparrow$  & SIR$\uparrow$ & SAR$\uparrow$ \\
        \midrule
        DAVIS-Flow (visual) & &{5.66} & {10.62} & {10.63} \\  
        DAVIS-Flow (text)   & &3.53 & 8.17 & 9.81 \\ 
        AMnet              & &3.71 & 9.15 & 11.00 \\
        \bottomrule
    \end{tabularx}
\end{table}

\Revision{
\subsection{Discussion: Extension to Complex-Valued Spectrogram Modeling}
In our current setting, we reconstruct audio using the predicted magnitude combined with the mixture phase.  
While effective, this approach can limit the fidelity of the separated audio, particularly in scenarios where spatial information is important.  
A promising extension is to adapt the existing framework to jointly model magnitude and phase.  
One intuitive way to achieve this is to transform the audio into the concatenation of the real and imaginary parts of the complex-valued spectrogram, replacing the current magnitude-only input.  
However, adopting this representation introduces practical challenges: 
\textit{(i) Preprocessing:} Real and imaginary components have different statistics and sign structures, requiring careful normalization to avoid bias. 
\textit{(ii) Scaling:} Both components must be scaled consistently (e.g., under log compression) to preserve phase coherence; mismatches can distort relative amplitudes. 
\textit{(iii) Frequency characteristics:} Variance patterns differ across frequencies, especially at high frequencies where phase noise is stronger, complicating learning and loss design. 
\textit{(iv) Model complexity:} Complex inputs double channel dimensionality and introduce sign-sensitive patterns, which may require specialized architectures, normalization, or phase-aware loss functions.

We leave the full exploration of robust complex-valued spectrogram modeling as an important avenue for future work.

}

\section{Conclusion}
In this paper, we propose DAVIS, a diffusion-based audio-visual separation framework designed to address the problem in a generative manner.  Unlike conventional discriminative methods, DAVIS is built upon a $T$-step diffusion model, enabling the iterative synthesis of the separated magnitude spectrogram conditioned on the visual input. By proposing a specialized Separation U-Net coupled with a novel sampling strategy, we successfully apply diffusion model to this new task, yielding high-quality sound separation results. Building on this framework, and leveraging Flow Matching, we develop DAVIS-Flow, a variant that demonstrates performance surpassing that of the vanilla DAVIS.  Extensive experiments on the MUSIC and AVE datasets validate both DAVIS's and DAVIS-Flow's effectiveness in separating sounds within specific and open domains, as well as their ability to deal with diverse time-frequency structures.
\backmatter

\section{Data Availability}

The publicly available datasets used in this work are:
\begin{itemize}
  \item AVE: Audio–Visual Event dataset \citep{tian2018audio}, available at \url{https://github.com/YapengTian/AVE-ECCV18}.
  \item MUSIC dataset~\citep{zhao2018sound}, available at \url{https://github.com/roudimit/MUSIC_dataset}.
\end{itemize}

Our code and pre‐trained models introduced in this paper are released under an MIT license at \url{https://github.com/WikiChao/DAVIS}.

If any additional data are needed, they are available from the corresponding author upon reasonable request.

\begin{appendices}

\section{Training and Inference Pseudo Code}
\label{subsec:algorithms}

The complete training and inference procedures for our DAVIS and DAVIS-Flow methods are shown in \cref{alg:unified_training} and \cref{alg:unified_inference}.
\begin{algorithm}[t!]
\caption{Training}
\label{alg:unified_training}
\begin{algorithmic}[1]
\State \textbf{Input:} A dataset $D = \{(a^{(k)}, v^{(k)})\}_{k=1}^K$
\State \textbf{Initialize:} randomly initialize $f_\theta$, $\phi(\cdot)$, load $\mathbf{Enc_v}$
\State \textbf{repeat}
\State \hskip1em Sample $(a^{(1)}, v^{(1)})$ and $(a^{(2)}, v^{(2)}) \sim D$
\State \hskip1em Mix and compute $x^{mix}, x^{(1)}$ 
\State \hskip1em Scale $x^{mix}, x^{(1)}$ using $\log(1+x) \cdot \sigma$ and clip to [0,1]
\State \hskip1em Encode $\boldsymbol{v^{(1)}} := \phi(\mathbf{Enc_v}(v^{(1)}))$
\State \hskip1em Sample $\epsilon \sim \mathcal{N}(\textbf{0, I})$
\If{method type is \texttt{DAVIS}}
\State \hskip1em Sample discrete $t \sim$ Uniform$ {(1,...,T)}$
\State \hskip1em Compute $x_t$ using DDPM forward process: $x_t = \sqrt{\Bar{\alpha}_t}x^{(1)} +\sqrt{1 - \Bar{\alpha}_t}\epsilon$
\State \hskip1em $\epsilon_t= f_\theta(x_t, x^{mix}, \boldsymbol{v}^{(1)}, t)$
\State \hskip1em Compute loss: $\mathcal{L}= ||\epsilon - \epsilon_t||_1$
\EndIf
\If{method type is \texttt{DAVIS-Flow}}
\State \hskip1em Sample continuous $t \sim \text{Uniform}( [0, 1])$
\State \hskip1em Compute $x_t = (1-t)\epsilon + tx^{(1)}$
\State \hskip1em $v_t = f_\theta(x_t, x^{mix}, \boldsymbol{v}^{(1)}, t)$
\State \hskip1em Define target vector field: $u_t = x^{(1)} - \epsilon$
\State \hskip1em Compute loss: $\mathcal{L}= ||v_t - u_t ||_1$
\EndIf
\State \hskip1em Take gradient step on $\nabla_\theta \mathcal{L}$
\State \textbf{until} converged
\end{algorithmic}
\end{algorithm}

\begin{algorithm}[t!]
\caption{Inference}
\label{alg:unified_inference}
\begin{algorithmic}[1]
\State \textbf{Input:} Audio mixture $a^{mix}$, visual frame $v$, number of sampling steps $N$, network $f_\theta$, $\phi(\cdot)$, $\mathbf{Enc_v}$
\State Compute $x^{mix} := \mathbf{STFT}(a^{mix})$
\State Encode $\boldsymbol{v^{(1)}} := \phi(\mathbf{Enc_v}(v))$
\If{method type is \texttt{DAVIS}}
\State Sample initial noise $x_{\text{cur}} \sim \mathcal{N}(\textbf{0, I})$
\For{$t = N, ..., 1$}
\State Sample $z \sim \mathcal{N}(\textbf{0, I})$ if $t>1$, else $z=0$
\State $\hat{\epsilon} = f_\theta(x_{\text{cur}}, x^{mix}, \boldsymbol{v^{(1)}}, t)$
\State Compute previous noisy sample: $x_{\text{prev}} = \frac{1}{\sqrt{\alpha_t}}(x_{\text{cur}} - \frac{1 - \alpha_t}{\sqrt{1 - \Bar{\alpha}_t}}\hat{\epsilon}) + \sqrt{\Tilde{\beta_t}}z$
\State $x_{\text{prev}} = \text{silence\_guided\_sampling}(x_{\text{prev}})$
\EndFor
\EndIf
\If{method type is \texttt{DAVIS-Flow}}
\State Sample initial noise $x_{\text{cur}} \sim \mathcal{N}(\textbf{0, I})$
\State Set time step $\Delta t = 1/N$
\For{$i = 1, ..., N$}
\State Set current time $t = i \cdot \Delta t$
\State $\hat{v} = f_\theta(x_{\text{cur}}, x^{mix}, \boldsymbol{v^{(1)}}, t)$
\State Update the sample (Euler step): $x_{\text{prev}} = x_{\text{cur}} + \hat{v} \cdot \Delta t$
\State Apply silence\_guided\_sampling to $x_{\text{prev}}$
\EndFor
\EndIf
\State \textbf{return} $e^{x_{prev}/\sigma} - 1$
\end{algorithmic}
\end{algorithm}

\section{More Qualitative Visualizations from Diverse Categories}
To demonstrate the effectiveness of our method in separating sounds across diverse categories, we conducted experiments on VGGSound~\citep{chen2020vggsound}.  VGGSound is a large-scale audio-visual dataset encompassing a wider range of sounds compared to commonly used datasets like AVE~\citep{tian2018audio} and MUSIC~\citep{zhao2018sound}.  However, since VGGSound isn't a standard benchmark for assessing audio-visual separation performance yet, we thus constructed two smaller datasets focusing on the two main subcategories within VGGSound: Animals and Vehicles.

We named these two datasets \textbf{VGGSound-Animal$10$} and\textbf{ VGGSound-Vehicle$10$}.  For each subcategory, we randomly selected 10 classes and sampled 100 videos for training and 10 videos for testing within each class. This resulted in a training/testing split of 1000/100 videos for each dataset, exceeding the scale of the frequently used MUSIC dataset (11 categories, 468/26 videos for training/testing). Therefore, we believe our custom datasets provide a meaningful evaluation for audio-visual separation tasks, and we leave the exploration of constructing a larger, more comprehensive dataset with additional categories and more comprehensive comparison for future work.

\begin{figure*}[!htbp]
    \centering
    \includegraphics[width=0.99\textwidth]{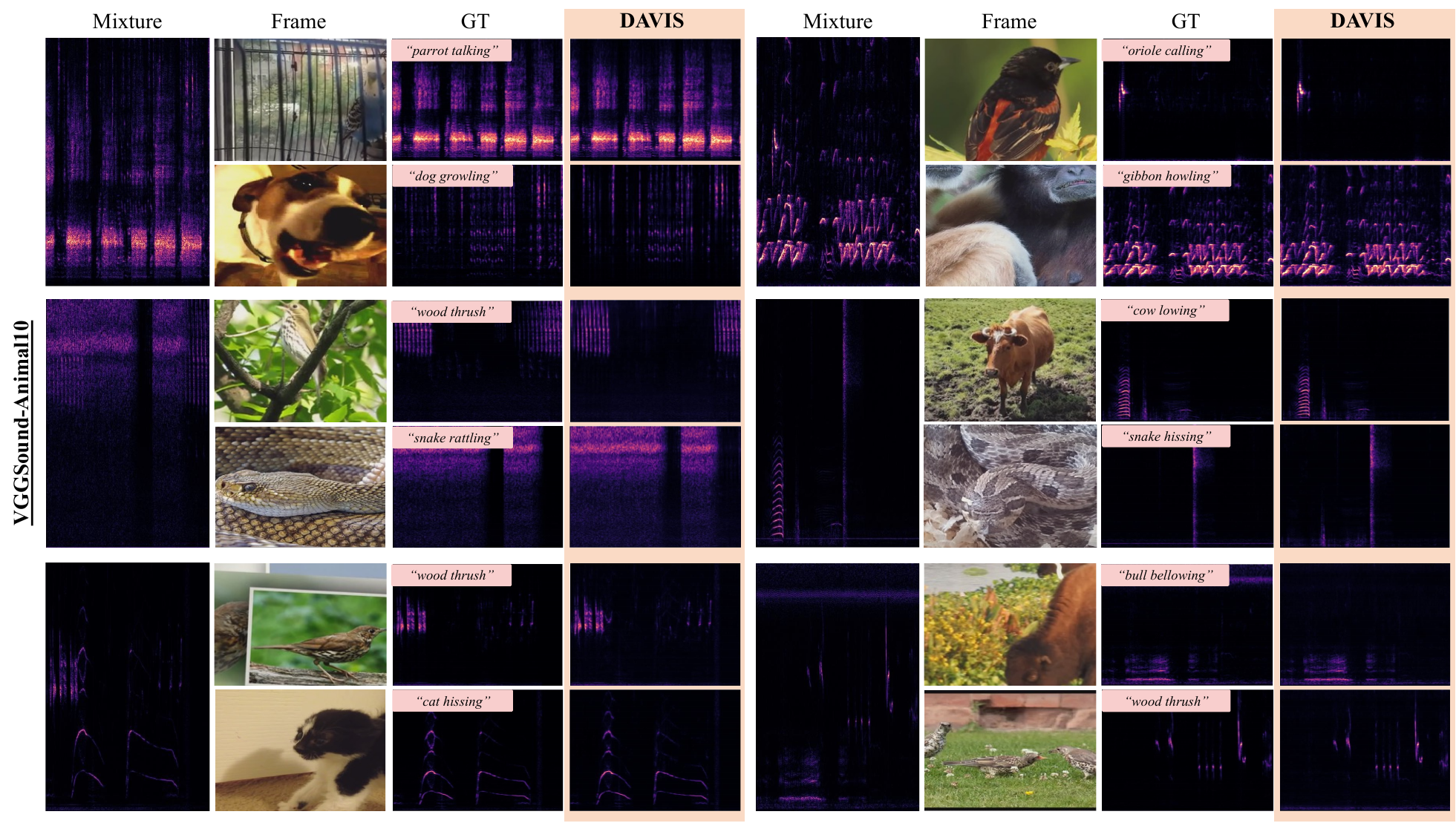}
    \caption{Qualitative visualizations on VGGSound-Animal$10$.}
    \label{fig:animal}
\end{figure*}

\begin{figure*}[!htbp]
    \centering
    \includegraphics[width=0.99\textwidth]{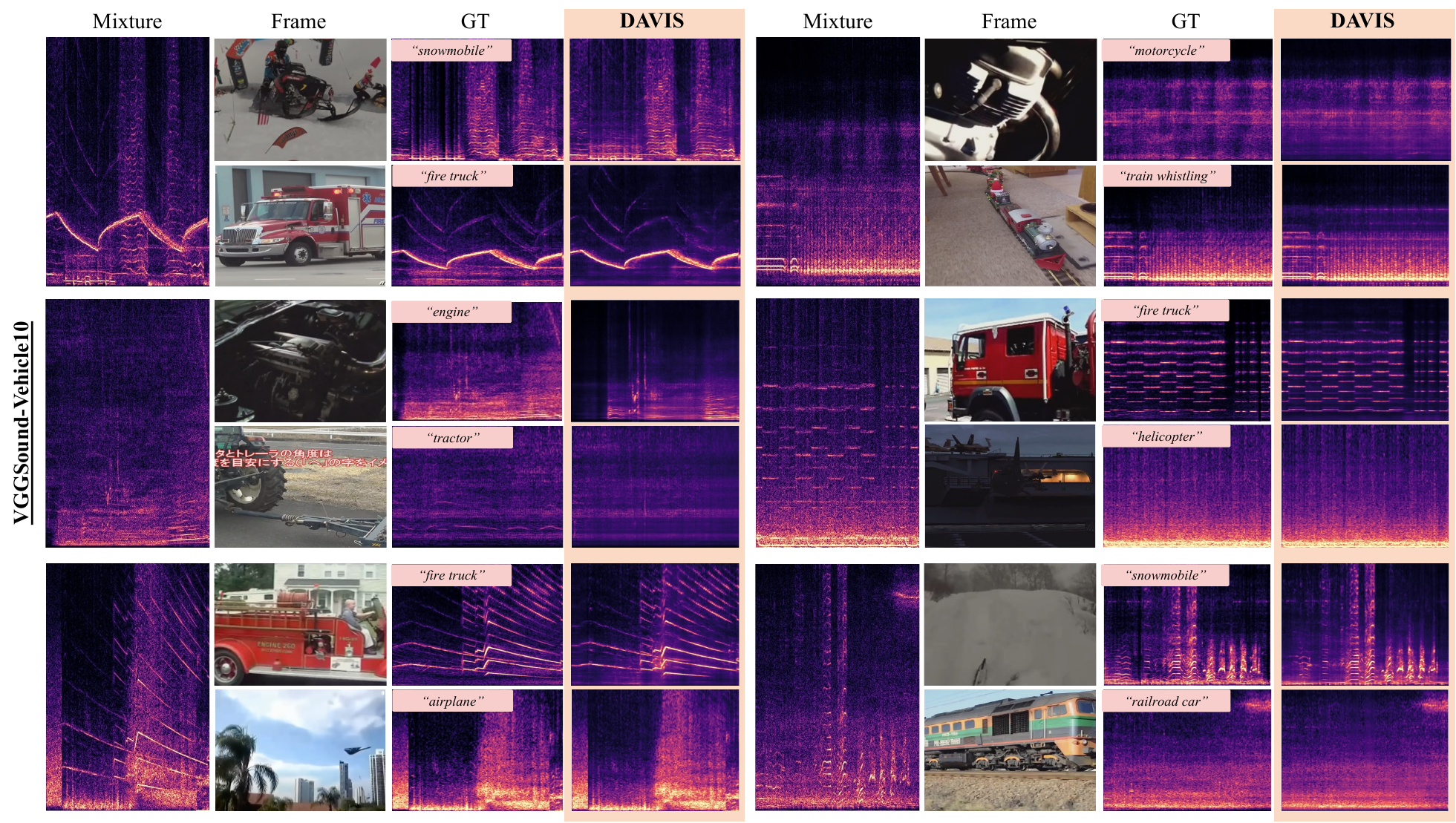}
    \caption{Qualitative visualizations on VGGSound-Vehicle$10$. There is a lot of background noise in this dataset, which makes it difficult to separate and poses significant challenges in achieving high-quality results.}
    \label{fig:vehicle}
\end{figure*}

The table in \cref{tab:vggsound} displays information about the classes in each dataset. We present qualitative results in \cref{fig:animal} and \cref{fig:vehicle}, which demonstrate that DAVIS can separate high-quality sounds from diverse categories.

\begin{table*}[!h]
    \centering
    \footnotesize
        \caption{Overview of our constructed \textbf{VGGSound-Animal$10$} and \textbf{VGGSound-Vehicle$10$} datasets. Each dataset contains 1000/100 videos for training/testing.}
    \begin{tabularx}{0.95\textwidth}{lX|l}
        \toprule                                              
        VGGSound-Animal10  & & VGGSound-Vehicle10  \\
        \midrule
        \textit{parrot talking} & & \textit{tractor digging}  \\
        \textit{dog growling} & & \textit{driving snowmobile}  \\
        \textit{cow lowing} & & \textit{reversing beeps}  \\
        \textit{cat hissing} & & \textit{helicopter}  \\
        \textit{gibbon howling} & & \textit{train whistling}  \\
        \textit{wood thrush calling} & & \textit{airplane flyby}  \\
        \textit{snake hissing} & & \textit{railroad car, train wagon}  \\
        \textit{baltimore oriole calling} & & \textit{driving motorcycle}  \\
        \textit{bull bellowing} & & \textit{engine accelerating, revving, vroom}  \\
        \textit{snake rattling} & & \textit{fire truck siren}  \\
        \bottomrule
    \end{tabularx}
    \label{tab:vggsound}
\end{table*}

\section{Implementation Details}
\label{sec:implementation}
In our experimental setup, we down-sample audio signals at 11kHz. For the MUSIC dataset, the video frame rate is set to 8 fps. Each video is approximately 6 seconds and we uniformly select 11 frames per video. The pre-trained ResNet18~\citep{he2016deep} is used as the image encoder. As for the AVE dataset, we set the video frame rate to 1 fps (following the setup of \citep{tian2018audio}). We use the entire 10-second audio as input and use 10 frames to train the model. The pre-trained CLIP image encoder~\citep{radford2021learning} is used.
During training, the frames are first resized to 256$\times$256 and then randomly cropped to $224\times224$. We set the total diffusion time step $T=1000$ to train our DAVIS model, and we uniformly sample flow matching time step from $t\in[0,1]$ for training our DAVIS-Flow model.
During inference, all the frames are directly resized to the desired size without cropping. To accelerate the separation process, we use DDIM~\citep{song2020denoising} with a sampling step of 15 for DAVIS, and adopt Euler Solver~\citep{lipman2022flow} with a sampling step of 2 for DAVIS-Flow.
The audio waveform is transformed into a spectrogram with a Hann window of size 1022 and a hop length of 256.
The obtained magnitude spectrogram is subsequently resampled to $256\times 256$ to feed into the separation network. We set the number of audio and visual feature channels $C$ as 512 and empirically choose the scale factor $\sigma=0.15$. 
Our model is trained with the Adam optimizer, with a learning rate of $10^{-4}$. The training is conducted on two 4090 GPUs for 200 epochs with a batch size of 8.

\section{Limitation}
\noindent\textbf{Visual Embeddings.} Our proposed DAVIS framework incorporates the extraction of global visual embedding as a condition for visually-guided source separation. This technique, which utilizes global visual features, has been widely adopted in audio-visual learning~\citep{zhao2018sound,huang2023egocentric}. Unlike methods that rely on pre-trained object detectors for extracting visual features, our framework does not have such a dependency. However, it may encounter limitations when trained on unconstrained video datasets. Intuitively, successful results can be achieved when the video contains a distinct sounding object, such as solo videos in the MUSIC dataset or videos capturing a sounding object performing a specific event in the AVE dataset.
Nonetheless, this training assumption may not hold in more challenging scenarios, where multiple objects are likely producing sounds, rendering the global visual embedding inadequate for accurately describing the content of sounding objects. To address this issue, one possible approach is to adapt our framework to leverage more fine-grained visual features and jointly learn sounding object localization and visually-guided sound separation. This adaptation enables the model to utilize localized sounding object information to enhance the audio-visual association.

\noindent\textbf{Evaluation Metrics.} We observe from the examples on the AVE and VGGSound datasets that many video clips contain off-screen sound or background noise, rendering the notion of ground truth unsuitable for evaluation. Consequently, comparing separation results with the ground truth audio clip and reporting SDR/SIR/SAR values may be insufficient to assess the method's effectiveness. Therefore, a new metric is needed to evaluate sound separation quality in more noisy or challenging scenarios. One possible metric is to leverage pretrained audio-text models and perform zero-shot classification or measure the cosine similarity between separated audio and the text label (analogous to zero-shot scenarios using the CLIP model).

\section{Future Work}
Our approach initiates the utilization of generative models for audio-visual scene understanding, paving the way for potential extensions to other multi-modal perception tasks like audio-visual object localization. Humans demonstrate the ability to imagine a ``dog'' upon hearing a ``barking'' sound, highlighting the potential of cross-modal generation in advancing audio-visual association learning. This implies that localization and separation tasks can be integrated into a single generative framework. 
In the future, we plan to explore the application of generative models to jointly address audio-visual localization and separation tasks.

\noindent\textbf{More Types of Conditions.} We have investigated the use of visual frame features and text prompts as conditions for sound separation in our work. These conditions are effective for separating sounds from different categories. However, for a more challenging scenario such as separating sounds from the same category, we need a different type of condition that provides discriminative cues to guide separation. Examples of such conditions are optical flow and trajectory. In our future work, we plan to incorporate more conditions in our framework.

\end{appendices}

\bibliographystyle{apalike}
\bibliography{sn-bibliography}%

\end{document}